\documentclass[]{imag-ms-template}

\usepackage[utf8]{inputenc} 
\usepackage[T1]{fontenc}  
\usepackage{csquotes}    
\usepackage[backend=biber, style=apa]{biblatex}

\usepackage{placeins}
\usepackage{caption}
\usepackage{longtable}
\usepackage{array}
\usepackage{tabularx}
\usepackage{makecell}
\usepackage{multirow}
\usepackage{float}
\usepackage{booktabs}
\usepackage{threeparttable}
\usepackage{amsmath}
\usepackage{amssymb}
\usepackage{graphicx}
\usepackage{url}
\usepackage{microtype}

\title{C-WFDM: A ControlNet-Guided Wavelet-Fusion Diffusion Model for Multimodal Brain MRI Completion}

\author{Muhammad Nabi Yasinzai,$^{1}$ Remika Mito,$^{2}$ Mangor Pedersen$^{1\ast}$\\
{\small $^{1}$Department of Psychology \& Neuroscience, Auckland University of Technology,}\\
{\small Auckland, New Zealand.}\\
{\small $^{2}$Department of Psychiatry, University of Melbourne, Melbourne, Victoria 3053, Australia.}\\
{\small $^\ast$Correspondence: mangor.pedersen@aut.ac.nz}
}

\begin{document}

\maketitle


\begin{abstract}

Multimodal brain MRI provides complementary anatomical, tissue, and pathological information to support downstream artificial intelligence (AI) applications. 
However, modality availability is uneven across public and pooled neuroimaging datasets: T1-weighted (T1w) MRI is widely represented, T2-weighted (T2w) and fluid-attenuated inversion recovery (FLAIR) MRI are less consistently available, and gadolinium-enhanced T1-weighted (T1Gd) MRI is relatively scarce.
Pooled datasets therefore contain participants with different combinations of acquired MRI contrasts, reflecting variation in the study designs, acquisition protocols, and clinical or research objectives of the contributing studies.
Many missing-modality benchmarks simulate incomplete MRI by withholding contrasts from complete four-modality cases, which may not represent the observed availability patterns in pooled datasets.
To address this gap, we introduce C-WFDM, a ControlNet extension of our Wavelet-Fusion Diffusion Model (WFDM), for subject-specific missing-modality completion.
C-WFDM is built on an earlier BrainScape snapshot comprising 157 public MRI datasets and 45{,}880 preprocessed scans.
The pretrained WFDM backbone retains target-modality and metadata conditioning, while target-specific ControlNet adapters use MRI contrasts from the same participant to guide synthesis.
C-WFDM synthesizes T2w from T1w, FLAIR from T1w and T2w, and T1Gd from T1w, T2w, and FLAIR.
Acquired conditioning contrasts are used when available, and generated intermediate contrasts are used only when required.
We compare C-WFDM with ALDM, cWDM, and Pix2Pix using paired structural similarity index (SSIM), peak signal-to-noise ratio (PSNR), and mean absolute error (MAE).
In the BrainScape completion experiment, C-WFDM achieves the best T2w and FLAIR results among the evaluated models, with SSIM values of $0.9232$ and $0.8904$, respectively.
T1Gd remains a challenging target, consistent with the original WFDM study.
C-WFDM supports subject-specific missing-modality completion in heterogeneous pooled MRI datasets and could increase the availability of complete multimodal MRI sets for downstream AI research.

\end{abstract}

\FloatBarrier
\section{Introduction}
\label{sec:introduction}

Multimodal structural magnetic resonance imaging (MRI) is widely used in neuroimaging analysis and clinical artificial intelligence (AI) because different contrasts provide complementary information about brain anatomy, tissue properties, and pathological features.
For example, brain tumor imaging commonly uses T1-weighted (T1w), T2-weighted (T2w), gadolinium-enhanced T1-weighted (T1Gd), and fluid-attenuated inversion recovery (FLAIR) MRI, which provide complementary information for identifying and delineating tumor subregions \parencite{Menze-BRATS-2014,Bakas-BRATS-2018}.
This dependence on multiple MRI contrasts makes the availability of MRI modalities an important consideration when developing and evaluating downstream AI models, because missing modalities can remove complementary anatomical and pathological information and reduce the reliability of downstream neuroimaging analyses \parencite{Kim-ALDM-2024,Meng-M2DN-2024}.

Complete multimodal MRI acquisition protocols can be clinically justified for selected patient groups, and acquiring the same set of modalities is often not practical or clinically appropriate for every participant.
This is particularly relevant to research cohorts and healthy controls, as MRI acquisition depends on diagnostic objectives, clinical feasibility, scan time, cost, patient tolerance, and safety considerations \parencite{Jiang-CoLaDiff-2023,Kim-ALDM-2024}.
This constraint is evident for T1Gd, which highlights abnormal contrast enhancement associated with blood-brain barrier disruption in clinically relevant pathologies, including brain tumors \parencite{Zhou-GadoliniumCancerImaging-2012}. 
Because gadolinium-based contrast agents are used only when clinically justified, T1Gd is common in selected clinical imaging protocols but rare in healthy or population cohorts.
Neuroimaging datasets therefore often have incomplete modality coverage, with many participants lacking a complete set of multimodal MRI sequences.

The missing-modality problem becomes more visible in large pooled neuroimaging resources. These resources combine data acquired for different scientific questions, clinical indications, participant populations, scanner hardware, and acquisition protocols, resulting in structured differences in modality availability \parencite{Yasinzai-BrainScape-2025}.
This pattern is evident in BrainScape, our previously developed open-source framework and curated resource for downloading, organizing, preprocessing, and integrating anatomical MRI datasets from multiple public repositories \parencite{Yasinzai-BrainScape-2025}.
BrainScape includes T1w, T2w, T1Gd, and FLAIR MRI across a large collection of public datasets, while preserving source provenance, acquisition information, and available demographic or clinical metadata \parencite{Yasinzai-BrainScape-2025}.
The BrainScape snapshot used in this study comprises 157 datasets and 45,880 preprocessed scans across T1w, T2w, FLAIR, and T1Gd, including 30,965 T1w, 6,893 T2w, 6,552 FLAIR, and 1,470 T1Gd scans.
BrainScape is a realistic and practically challenging setting for multimodal MRI completion because its modality availability reflects real-world heterogeneity across pooled neuroimaging datasets.

In our previous work, we introduced the Wavelet-Fusion Diffusion Model (WFDM) for multimodal brain MRI synthesis \parencite{Yasinzai-WFDM-2026}.
WFDM generates T1w, T2w, FLAIR, and T1Gd MRI by conditioning the model on the target modality along with demographic and clinical metadata available in BrainScape \parencite{Yasinzai-WFDM-2026}.
WFDM combines the Wavelet-Fusion variational autoencoder (WF-VAE) with a conditional 3D U-Net diffusion model trained in the WF-VAE latent space using modality and metadata conditioning.
The WF-VAE compresses 3D MRI volumes into a compact wavelet-fusion latent space. The diffusion model learns to generate samples in this latent space, and the WF-VAE decoder maps generated latents back to 3D MRI volumes.
By performing diffusion in the compressed latent space, WFDM reduces the memory and computational burden associated with full-resolution 3D diffusion \parencite{Yasinzai-WFDM-2026,Pinaya-LDM3DBrain-2022}.
However, WFDM is a cohort-level modality generator rather than a subject-specific completion model.
Although it can generate MRI volumes across T1w, T2w, FLAIR, and T1Gd using target-modality and metadata conditioning, it does not use partially available scans from the same participant to synthesize missing modalities.
This limits its direct use in pooled datasets where each participant may have only a subset of modalities available, and the goal is to complete the multimodal MRI set for that participant.

Many studies synthesize missing MRI modalities from the available modalities of the same participant.
Generative adversarial network (GAN) based image-to-image translation methods, such as Pix2Pix, provide a non-diffusion baseline for supervised paired translation \parencite{Isola-Pix2Pix-2017}.
More recent studies have also explored diffusion-based architectures for MRI modality completion.
Among these diffusion-based methods, ALDM uses latent diffusion to translate a single source MRI contrast into a target contrast using target-modality conditioning \parencite{Kim-ALDM-2024}.
Similarly, cWDM synthesizes one missing modality from three acquired MRI modalities and performs diffusion directly in a wavelet representation \parencite{Friedrich-cWDM-2024}.
Other studies have explored cycle-guided cross-modality synthesis and volume-aware generation strategies \parencite{Pan-CGDDPM-2025,Zhu-MakeAVolume-2023}.
Additional methods have explored flexible completion strategies using modality masks, adaptive image-frequency fusion, feature mapping and merging, frequency-guided diffusion, and multiscale structure-guided latent diffusion \parencite{Meng-M2DN-2024,Kebaili-AMMDiff-2025,Zhong-FMMDiff-2025,Xiao-FgC2FUDiff-2024,Lin-MSGLDM-2026}.
Many of these methods are evaluated on complete multimodal datasets such as BraTS, where T1w, T1Gd, T2w, and FLAIR are available for each participant.
In these evaluations, missing modalities are often simulated by withholding acquired contrasts or applying modality masks to otherwise complete cases.
Although such protocols are valuable for controlled benchmarking, they do not fully represent large real-world pooled neuroimaging datasets, where modality availability is sparse, uneven, and shaped by dataset source, acquisition protocol, clinical indication, and the selective use of contrast-enhanced imaging \parencite{Yasinzai-BrainScape-2025}.

A modality-completion framework for large pooled MRI resources should follow the empirical modality availability patterns of the target dataset.
Instead of treating all translation directions as equally likely or clinically meaningful, completion should start from the most consistently available modality and then target less frequently available modalities.
Modality completion should therefore not be formulated as a symmetric paired-translation problem, because some directions are clinically uncommon and poorly supported by real modality-availability patterns.
For example, completing T2w, FLAIR, or T1Gd from T1w is better aligned with pooled-dataset acquisition patterns. 
In contrast, generating T1w from T1Gd is poorly aligned with real modality availability because T1Gd is clinically selective, far less frequent, and rarely available as the only useful source contrast.

We introduce C-WFDM, a ControlNet extension of WFDM for subject-specific multimodal brain MRI completion. 
C-WFDM retains the pretrained WFDM backbone with target-modality and metadata conditioning and introduces target-specific ControlNet adapters that incorporate anatomical information from available MRI contrasts of the same participant. 
The pretrained WFDM components remain fixed while the ControlNet adapters are trained. 
C-WFDM follows a sequential modality-completion policy motivated by the modality-availability patterns observed in BrainScape. 
The model first synthesizes T2w from T1w, then synthesizes FLAIR from T1w and T2w, and finally synthesizes T1Gd from T1w, T2w, and FLAIR.
Real conditioning modalities are used whenever available, while synthetic intermediate modalities are used only when required.

We evaluate C-WFDM in three complementary experiments. 
The first evaluates BrainScape completion under observed modality-availability patterns.
The second evaluates controlled BraTS completion using acquired conditioning contrasts. 
The third examines BraTS T1w-only sequential completion, in which T2w, FLAIR, and T1Gd are synthesized successively using a single acquired T1w scan and generated intermediate contrasts.  
We also measure inference time and peak GPU memory usage to assess computational efficiency.

\FloatBarrier
\section{Methodology}
\label{sec:methodology}

\subsection{Ethics statement}
\label{subsec:ethics}

This study is a secondary analysis of de-identified anatomical MRI data obtained from publicly available source datasets integrated through BrainScape. 
No new human data were collected for the present work. 
Ethical approval, participant consent, and requirements for data access are governed by the original contributing studies and repositories from which the BrainScape datasets were derived.

\subsection{Dataset}
\label{subsec:dataset}

\subsubsection{BrainScape overview}
\label{subsubsec:brainscape-overview}

C-WFDM training uses the earlier fixed BrainScape snapshot and subject-aware splits from our WFDM study \parencite{Yasinzai-WFDM-2026}.
The fully published BrainScape resource aggregates 160 publicly available anatomical MRI datasets comprising 27{,}227 participants and 46{,}583 scans after quality control.
BrainScape contains the four anatomical MRI contrasts used in this study, including T1w, T2w, T1Gd, and FLAIR, and retains demographic, clinical, and acquisition metadata whenever these fields are available in the originating datasets \parencite{Yasinzai-BrainScape-2025}.

The experimental snapshot used here comprises 157 datasets and 45{,}880 preprocessed MRI scans across T1w, T2w, T1Gd, and FLAIR \parencite{Yasinzai-WFDM-2026}.
This snapshot was finalized before three additional datasets were incorporated into the published BrainScape release.
The modality counts reported in Table~\ref{tab:brainscape-splits} therefore differ slightly from the published BrainScape statistics.
The experimental snapshot contains fewer T1w and T2w scans, whereas the T1Gd and FLAIR counts are unchanged relative to the published BrainScape release.

\begin{table}[t]
\centering
\begin{threeparttable}
\caption{BrainScape experimental snapshot used for WFDM and C-WFDM training, validation, and testing.}
\label{tab:brainscape-splits}
\setlength{\tabcolsep}{14pt}
\renewcommand{\arraystretch}{1.1}
\begin{tabular}{@{}lrrrr@{}}
\toprule
Modality & Total & Train & Val & Test \\
\midrule
T1w   & 30{,}965 & 24{,}515 & 3{,}036 & 3{,}414 \\
T2w   & 6{,}893  & 5{,}332  & 694     & 867     \\
FLAIR & 6{,}552  & 5{,}064  & 673     & 815     \\
T1Gd  & 1{,}470  & 1{,}001  & 172     & 297     \\
\midrule
MRI volumes & 45{,}880 & 35{,}912 & 4{,}575 & 5{,}393 \\
\bottomrule
\end{tabular}
\begin{tablenotes}[flushleft]\footnotesize
\item The counts correspond to the earlier fixed BrainScape snapshot used for the WFDM and C-WFDM experiments.
\end{tablenotes}
\end{threeparttable}
\end{table}

\subsubsection{BrainScape modality availability}
\label{subsubsec:availability}
The BrainScape dataset used in this study exhibits a strong imbalance in MRI modality availability.
Table~\ref{tab:modality-combinations} summarizes the observed MRI modality combinations in BrainScape.
The table contains 31{,}430 dataset entries and is distinct from the scan-level BrainScape total reported in Table~\ref{tab:brainscape-splits}.
Percentages are computed relative to the 31{,}430 combination entries represented in this split table.
The distribution shows that 20{,}549 dataset entries contain only T1w, corresponding to 65.38\% of the table.
The next most common combinations are T1w with T2w and T1w with FLAIR, corresponding to 13.57\% and 11.43\% of dataset entries, respectively.
In total, 30{,}965 entries contain T1w, corresponding to 98.52\% of the table. The remaining 465 entries do not contain T1w, corresponding to 1.48\% of the table.
These counts motivate the choice of T1w as the starting contrast for the C-WFDM modality-completion pathway.
Among the 30{,}965 T1w-containing entries, 6{,}824 also contain T2w and 6{,}152 also contain FLAIR, corresponding to 22.04\% and 19.87\% of T1w-containing entries, respectively.
Complete four-modality coverage is available in only 1{,}470 dataset entries, corresponding to 4.68\% of the table.
The BrainScape completion problem is therefore asymmetric. 
T1w is widely available, T2w and FLAIR are less consistently available, and T1Gd is the least available modality.
The C-WFDM completion policy follows this observed structure by synthesizing T2w from T1w, FLAIR from T1w and T2w, and T1Gd from T1w, T2w, and FLAIR.
Real conditioning inputs are used whenever available, while synthetic intermediate modalities are used only when required.
Entries without T1w are retained in Table~\ref{tab:modality-combinations} to describe the observed modality combination patterns, but they are not used by the sequential completion pathway.

\begin{table*}[t]
\centering
\begin{threeparttable}
\caption{MRI modality combinations in the BrainScape experiment snapshot.}
\label{tab:modality-combinations}
\newcommand{\modpresent}{\ensuremath{\bullet}}
\newcommand{\modabsent}{\ensuremath{\circ}}
\setlength{\tabcolsep}{7pt}
\renewcommand{\arraystretch}{1.12}
\begin{tabular}{@{}ccccrrrrr@{}}
\toprule
T1w & T2w & FLAIR & T1Gd & Total & Train & Val & Test & Percent \\
\midrule
\modpresent & \modabsent  & \modabsent  & \modabsent  & 20{,}549 & 16{,}373 & 1{,}996 & 2{,}180 & 65.38\% \\
\modpresent & \modpresent & \modabsent  & \modabsent  & 4{,}264  & 3{,}402  & 407     & 455     & 13.57\% \\
\modpresent & \modabsent  & \modpresent & \modabsent  & 3{,}592  & 2{,}868  & 352     & 372     & 11.43\% \\
\modpresent & \modpresent & \modpresent & \modpresent & 1{,}470  & 1{,}001  & 172     & 297     & 4.68\% \\
\modpresent & \modpresent & \modpresent & \modabsent  & 1{,}090  & 871      & 109     & 110     & 3.47\% \\
\modabsent  & \modabsent  & \modpresent & \modabsent  & 396      & 320      & 40      & 36      & 1.26\% \\
\modabsent  & \modpresent & \modabsent  & \modabsent  & 65       & 54       & 6       & 5       & 0.21\% \\
\modabsent  & \modpresent & \modpresent & \modabsent  & 4        & 4        & 0       & 0       & 0.01\% \\
\midrule
-- & -- & -- & -- & 31{,}430 & 24{,}893 & 3{,}082 & 3{,}455 & 100.00\% \\
\bottomrule
\end{tabular}
\begin{tablenotes}[flushleft]\footnotesize
\item[] \textit{Note.} Filled and open circles indicate present and absent MRI modalities, respectively. Percentages are computed relative to 31{,}430 BrainScape dataset entries.
\end{tablenotes}
\end{threeparttable}
\end{table*}

\subsection{Preprocessing}
\label{subsec:preprocessing}

The heterogeneous source datasets are preprocessed using the BrainScape BraTS preprocessing pipeline before model training \parencite{Yasinzai-BrainScape-2025}.
The pipeline aligns modalities within each participant, standardizes volumes to atlas space, performs brain extraction and intensity normalization, and resamples all volumes to a uniform resolution.

When multiple modalities are available for a given entry, intra-subject co-registration is performed to a single reference modality selected according to the following priority order:
$\mathrm{T1w} > \mathrm{T2w} > \mathrm{T1Gd} > \mathrm{FLAIR}$.
This step aligns available modalities within each participant and reduces contrast misalignment before model training.
Following co-registration, all volumes are rigidly registered to the SRI-24 atlas for consistent spatial alignment and orientation \parencite{Yasinzai-BrainScape-2025}.
Brain extraction is then performed using HD-BET to generate a brain mask, which is used to remove non-brain regions from each MRI volume.
After registration and masking, the BrainScape BraTS pipeline applies intensity normalization and resamples all volumes to isotropic $1\,\mathrm{mm}$ spacing \parencite{Yasinzai-BrainScape-2025}.

For model training, volumes are represented in RAS orientation as channel-first tensors and are cropped or padded to a fixed spatial size of $(H,W,D)=(192,224,160)$, yielding a tensor shape $(C,H,W,D)=(1,192,224,160)$.
In the PyTorch data module, each volume is additionally normalized using min-max scaling with high-percentile clipping before intensities are mapped to the range $[0,1]$.

\subsection{C-WFDM method overview}
\label{subsec:cwfdm-method-overview}

C-WFDM extends the WFDM generator from cohort-level target-contrast synthesis to subject-specific missing-modality completion.
It reuses the WF-VAE, the conditional WFDM denoising model, and the WFDM modality and metadata conditioning.
C-WFDM adds target-specific ControlNet adapters to WFDM, so target synthesis also uses the available modalities from the same participant for conditioning.
ControlNet conditioning is performed in WF-VAE latent space. The completion policy starts with T1w and sequentially synthesizes T2w from T1w, FLAIR from T1w and T2w, and T1Gd from T1w, T2w, and FLAIR.
At inference time, acquired contrasts are used for ControlNet conditioning whenever available, while synthetic intermediates are used only when the corresponding real conditioning contrast is missing.

\subsection{Model architecture}
\label{subsec:model-architecture}

\begin{figure*}[p]
\centering
\includegraphics[width=\textwidth,height=0.95\textheight,keepaspectratio]{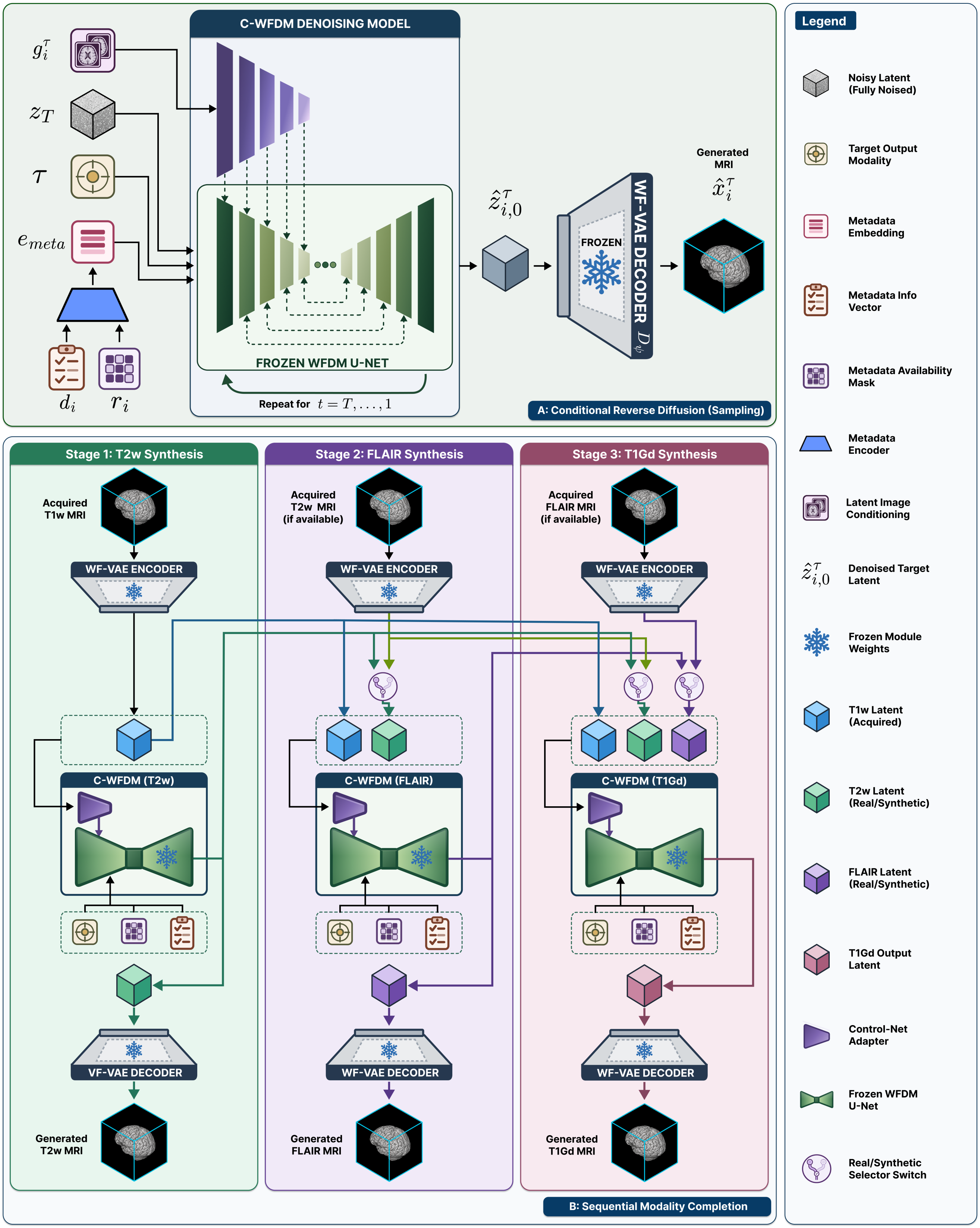}
\caption{Overview of the C-WFDM architecture.
(A) A trainable, target-specific ControlNet adapter provides latent image conditioning to the frozen WFDM denoising U-Net.
(B) The sequential completion pipeline starts with acquired T1w and completes T2w, FLAIR, and T1Gd in sequence, using available MRI scans or previously generated outputs for conditioning.}
\label{fig:cwfdm-architecture}
\end{figure*}

The C-WFDM architecture consists of the pretrained WFDM backbone and target-specific ControlNet adapters (Figure~\ref{fig:cwfdm-architecture}A).
The WFDM study supplies the pretrained WF-VAE encoder and decoder, the conditional denoising model for latent-space multimodal MRI generation, and the corresponding target-modality and metadata conditioning.
The ControlNet adapters use available modalities from the same participant as anatomical conditioning input when synthesizing a missing modality.

\subsubsection{WFDM backbone}
\label{subsubsec:model-wfdm-backbone}
C-WFDM uses the pretrained WFDM model as its latent generative backbone.
WFDM combines a Wavelet-Fusion VAE compression model with a conditional 3D U-Net denoising network trained in WF-VAE latent space \parencite{Yasinzai-WFDM-2026}.
The WF-VAE encoder compresses MRI volumes into latent space, and the WF-VAE decoder reconstructs MRI volumes from the generated latent representations.
The WFDM denoising network synthesizes target latents conditioned on the requested modality and the available demographic and clinical metadata.

\subsubsection{ControlNet-guided subject-specific conditioning}
\label{subsubsec:model-controlnet-adapter}

C-WFDM adds target-specific ControlNet adapters to the pretrained WFDM denoising model.
The adapters use WF-VAE latent representations of the participant's available modalities as image-conditioning inputs.
For each target modality, the selected conditioning latents follow the ordered modality tuple defined in Section~\ref{subsubsec:cwfdm-conditioning}.
The conditioning latents are concatenated channel-wise and passed to the target-specific ControlNet adapter.
At each denoising step, the ControlNet adapter receives the current noisy target latent, the diffusion timestep, and the latent conditioning tensor, and produces control features for the WFDM denoising U-Net.
The original WFDM target-modality and metadata conditioning remains unchanged.
Separate adapters are used for T2w, FLAIR, and T1Gd because each target uses a different set of conditioning modalities.

\subsection{Problem formulation}
\label{subsec:problem}

C-WFDM builds on WFDM, which performs multimodal MRI synthesis using target-modality and metadata conditioning.
In WFDM, generation is conditioned on the requested target modality, encoded demographic or clinical metadata, and the corresponding metadata availability mask \parencite{Yasinzai-WFDM-2026}.
C-WFDM retains the pretrained WFDM conditioning and adds subject-specific latent MRI conditioning from the available modalities of the same dataset entry.

\subsubsection{Multimodal dataset samples}
\label{subsubsec:multimodal-dataset-record}

Let $\mathcal{M}$ denote the set of MRI modalities considered in this study.
\[
\mathcal{M}=\left\{\mathrm{T1w},\mathrm{T2w},\mathrm{FLAIR},\mathrm{T1Gd}\right\}.
\]
For each dataset entry $i$ and modality $m\in\mathcal{M}$, let $x_i^m\in\mathbb{R}^{H\times W\times D}$ denote the preprocessed 3D MRI volume for that contrast when it is available.
All volumes use the WFDM preprocessing size $(H,W,D)=(192,224,160)$.
Availability of contrast $m$ for dataset entry $i$ is encoded by
\begin{equation}
\label{eq:modality-availability-indicator}
a_i^m\in\left\{0,1\right\}.
\end{equation}
Using this availability indicator, the observed modality set for dataset entry $i$ is
\[
\mathcal{A}_i=\left\{m\in\mathcal{M}\mid a_i^m=1\right\}.
\]
The corresponding set of observed MRI contrasts for dataset entry $i$ is
\[
\mathcal{X}_i=\left\{x_i^m\mid m\in\mathcal{A}_i\right\}.
\]
Each dataset entry may also include an encoded metadata vector $d_i$ and a metadata availability mask $r_i$, following the WFDM metadata-conditioning notation \parencite{Yasinzai-WFDM-2026}.
For a requested target modality $\tau\in\mathcal{M}$, the inherited WFDM conditioning object is
\begin{equation}
\label{eq:wfdm-conditioning-object}
c_{i,\tau}^{\mathrm{WFDM}}=(\tau,d_i,r_i).
\end{equation}

\subsubsection{Wavelet-Fusion VAE latent encoding}
\label{subsubsec:wfvae-latent-encoding}

Let $E_{\phi}$ denote the WF-VAE encoder described in Section~\ref{subsubsec:model-wfdm-backbone}.
For each dataset entry $i$ and available contrast $m\in\mathcal{A}_i$, the encoder outputs the mean $\mu_i^m$ and standard deviation $\sigma_i^m$ of the approximate posterior.
\[
\left(\mu_i^m,\sigma_i^m\right)=E_{\phi}\left(x_i^m\right).
\]
These tensors parameterize the diagonal Gaussian approximate posterior
\[
q_{\phi}\left(z_i^m\mid x_i^m\right)
=
\mathcal{N}\left(\mu_i^m,\operatorname{diag}\left((\sigma_i^m)^2\right)\right).
\]
For deterministic latent construction, the unscaled VAE latent is set to the posterior mean.
\[
\tilde{z}_i^m=\mu_i^m.
\]

This deterministic posterior-mean latent is used during inference by default.
During training, C-WFDM can instead construct $\tilde{z}_i^m$ from a posterior sample using the scheduled sampling rule described in Section~\ref{subsubsec:training-posterior-sampled-latents}.
With the WF-VAE downsampling factor $f=4$ and four latent channels, the unscaled latent has shape
\[
\tilde{z}_i^m\in\mathbb{R}^{4\times H/4\times W/4\times D/4}
=
\mathbb{R}^{4\times 48\times 56\times 40}.
\]

\subsubsection{WFDM latent scaling}
\label{subsubsec:wfdm-latent-scaling}

C-WFDM uses the same latent scaling convention as WFDM.
Each unscaled WF-VAE latent is multiplied by a fixed scalar $s_z$, estimated from the WFDM training split.
For any encoded MRI modality $m\in\mathcal{M}$, the scaled latent is
\[
z_i^m=s_z\tilde{z}_i^m.
\]
After denoising, the generated target latent is divided by the same scaling factor to recover the unscaled WF-VAE latent before decoding.
\[
\hat{x}_i^\tau=D_{\psi}\left(\frac{\hat{z}_{i,0}^\tau}{s_z}\right),
\]
where $\hat{z}_{i,0}^{\tau}$ denotes the final clean scaled target latent produced at the end of the reverse denoising process.

\subsubsection{Sequential completion policy}
\label{subsubsec:sequential-completion-policy}

The C-WFDM modality completion policy is based on the BrainScape dataset modality combinations summarized in Table~\ref{tab:modality-combinations}.
Because T1w is available for nearly all entries in the dataset, C-WFDM uses T1w as the starting contrast for the sequential completion.
Dataset entries without an acquired T1w scan are excluded from this sequential completion protocol. 
Missing modalities are completed in the fixed order
\[
\mathcal{T}=\left(\mathrm{T2w},\mathrm{FLAIR},\mathrm{T1Gd}\right).
\]
The corresponding completion sequence is
\[
\begin{aligned}
\mathrm{T1w} &\rightarrow \mathrm{T2w},\\
\mathrm{T1w}+\mathrm{T2w} &\rightarrow \mathrm{FLAIR},\\
\mathrm{T1w}+\mathrm{T2w}+\mathrm{FLAIR} &\rightarrow \mathrm{T1Gd}.
\end{aligned}
\]
Later stages can therefore condition on acquired T1w together with real or generated intermediate modalities from earlier completion steps (Figure~\ref{fig:cwfdm-architecture}B).
During inference, acquired upstream modalities are used whenever they are available.
If a required upstream modality is missing, C-WFDM uses the synthetic modality produced by the corresponding earlier completion step.
During training, when both acquired and synthetic intermediate latents are available, C-WFDM samples between them using the scheduled source-selection rule in Section~\ref{subsubsec:training-real-synthetic-scheduling} to reduce sensitivity to generated intermediates during sequential inference.

\subsubsection{C-WFDM conditioning}
\label{subsubsec:cwfdm-conditioning}

C-WFDM extends the WFDM conditioning object in Equation~\ref{eq:wfdm-conditioning-object} with subject-specific MRI latent conditioning for the ControlNet branch.
For each target modality $\tau\in\mathcal{T}$, let $\mathcal{C}_{\tau}$ denote the ordered tuple of modalities used for ControlNet conditioning.
\[
\begin{aligned}
\mathcal{C}_{\mathrm{T2w}} &= \left(\mathrm{T1w}\right),\\
\mathcal{C}_{\mathrm{FLAIR}} &= \left(\mathrm{T1w},\mathrm{T2w}\right),\\
\mathcal{C}_{\mathrm{T1Gd}} &= \left(\mathrm{T1w},\mathrm{T2w},\mathrm{FLAIR}\right).
\end{aligned}
\]

For each dataset entry $i$, target contrast $\tau$, and conditioning modality $m\in\mathcal{C}_{\tau}$, the available source set satisfies
\[
\mathcal{S}_{i,\tau}^m\subseteq\{\mathrm{R},\mathrm{S}\}.
\]
Here, $\mathrm{R}$ denotes a latent derived from an acquired MRI volume and $\mathrm{S}$ denotes the corresponding synthetic latent generated by an earlier C-WFDM completion step.
The source set is then
\[
\mathcal{S}_{i,\tau}^m
=
\begin{cases}
\left\{\mathrm{R}\right\},
& m=\mathrm{T1w},\\
\left\{\mathrm{S}\right\},
& m\in\mathcal{C}_{\tau},\ m\neq\mathrm{T1w},\ a_i^m=0,\\
\left\{\mathrm{R},\mathrm{S}\right\},
& m\in\mathcal{C}_{\tau},\ m\neq\mathrm{T1w},\ a_i^m=1.
\end{cases}
\]

The T1w conditioning modality is fixed to the acquired scan, while intermediate upstream modalities such as T2w and FLAIR may be represented by acquired or generated latents depending on availability and training policy.

Let $\rho_{i,\tau}^m\in\mathcal{S}_{i,\tau}^m$ denote the selected conditioning source.
At inference time, the selector prefers acquired upstream modalities whenever available and selects synthetic modalities only when the acquired inputs are missing.
During training, if $\mathcal{S}_{i,\tau}^m=\{\mathrm{R},\mathrm{S}\}$, either source may be sampled according to the scheduled source selection policy in Section~\ref{subsubsec:training-real-synthetic-scheduling}.
Given the selected source $\rho_{i,\tau}^{m}$, the posterior parameters used to construct the conditioning latent are
\[
(\mu_{i,\tau}^{m,\rho},\sigma_{i,\tau}^{m,\rho})
=
\begin{cases}
(\mu_i^m,\sigma_i^m), & \rho_{i,\tau}^m=\mathrm{R},\\
(\hat{\mu}_i^m,\hat{\sigma}_i^m), & \rho_{i,\tau}^m=\mathrm{S}.
\end{cases}
\]
The selected unscaled conditioning latent $\tilde{z}_{i,\tau}^{m}$ is then constructed from these selected posterior parameters as explained in Section~\ref{subsubsec:training-posterior-sampled-latents}.
The latent ControlNet conditioning tensor for target $\tau$ is formed by channel-wise concatenation of the selected scaled upstream latents in the order specified by $\mathcal{C}_{\tau}$.
\[
g_i^\tau
=
\mathop{\Vert}\limits_{m\in\mathcal{C}_{\tau}}
\left(
s_z\tilde{z}_{i,\tau}^{m}
\right).
\]
The full C-WFDM conditioning object is
\[
c_{i,\tau}^{\mathrm{C\mbox{-}WFDM}}
=
\left(
 c_{i,\tau}^{\mathrm{WFDM}},
 g_i^\tau
\right).
\]
C-WFDM keeps the WFDM target-modality and metadata conditioning and adds the latent ControlNet conditioning tensor $g_i^\tau$ (Figure~\ref{fig:cwfdm-architecture}A).

\subsection{Training protocol}
\label{subsec:training-protocol}

During training, only the ControlNet adapter parameters are optimized, while the WF-VAE compression model, the WFDM denoising U-Net, and the metadata encoder are kept fixed.
The adapter learns from the subject-specific latent conditioning input $g_i^\tau$ while the pretrained WFDM latent generator remains fixed.

For each target modality, C-WFDM trains a separate ControlNet adapter using paired dataset entries in which the target modality was acquired.
\[
\mathcal{I}_{\tau}^{\mathrm{train}}
=
\left\{
i\mid
a_i^{\mathrm{T1w}}=1,
a_i^\tau=1
\right\}.
\]
The target-availability condition is required because supervised adapter training needs an acquired reference target.
For each paired training example, the acquired target image is encoded into the WF-VAE latent space, scaled by the WFDM latent scaling factor, and noised or interpolated using the same scheduler used by the WFDM backbone.
The target-specific ControlNet adapter learns subject-specific features from $g_i^\tau$.

\subsubsection{Real and synthetic intermediate scheduling}
\label{subsubsec:training-real-synthetic-scheduling}

During supervised training, later stages may condition on real acquired intermediate modalities.
In contrast, sequential inference may require those same modalities to be generated by earlier C-WFDM stages when they are missing for a participant.

C-WFDM trains the ControlNet adapters using a mixture of acquired and synthetic intermediate conditioning latents.
This scheduling is applied only to intermediate conditioning modalities for which both real and synthetic latents are available.
\[
m\in\mathcal{C}_{\tau},
\qquad
m\neq\mathrm{T1w},
\qquad
\mathcal{S}_{i,\tau}^m=\left\{\mathrm{R},\mathrm{S}\right\}.
\]
For these cases, let $\pi_{e,\tau}^m\in[0,1]$ denote the epoch-dependent probability of selecting the acquired source for upstream modality $m$ when training target stage $\tau$ at epoch $e$.
The selected source is sampled as
\[
\Pr\left(\rho_{i,\tau}^m=\mathrm{R}\right)=\pi_{e,\tau}^m,
\qquad
\Pr\left(\rho_{i,\tau}^m=\mathrm{S}\right)=1-\pi_{e,\tau}^m.
\]
T1w always uses the acquired latent, while missing intermediate modalities use the generated latent from an earlier completion stage.
In the training schedule, generated intermediate latents are introduced progressively to improve the model's robustness to synthetic intermediate conditioning inputs.

\subsubsection{Posterior-sampled conditioning latents}
\label{subsubsec:training-posterior-sampled-latents}

For each selected real or synthetic conditioning modality, C-WFDM has access to the posterior mean latent and the corresponding posterior standard-deviation tensor.
The posterior mean is used by default, while the standard deviation is used only when posterior sampling is enabled during training.
For a selected source $\rho_{i,\tau}^{m}$ of conditioning modality $m$, the unscaled conditioning latent is
\[
\tilde{z}_{i,\tau}^m
=
\mu_{i,\tau}^{m,\rho}
+
b_{i,\tau}^m\alpha_e\sigma_{i,\tau}^{m,\rho}\odot\epsilon_{i,\tau}^m,
\]
where
\[
\epsilon_{i,\tau}^m\sim\mathcal{N}(0,I),
\qquad
b_{i,\tau}^m\sim\mathrm{Bernoulli}(p_e).
\]
Here, $p_e$ is the epoch-dependent probability of posterior sampling and $\alpha_e$ is the epoch-dependent posterior-sampling strength.
When $p_e=0$ or $\alpha_e=0$, the latent reduces to the selected posterior mean, $\tilde{z}_{i,\tau}^m=\mu_{i,\tau}^{m,\rho}$.
When posterior sampling is active, the selected posterior standard deviation and $\alpha_e$ together control the amount of latent variation introduced during training.

\subsubsection{Sequential training stages}
\label{subsubsec:training-sequential-stages}

Training is performed in stages, following the order of completion defined in Section~\ref{subsubsec:sequential-completion-policy}.
Each stage uses acquired T1w and target scans for supervised training, while generated intermediate conditioning inputs are used as described in Section~\ref{subsubsec:training-real-synthetic-scheduling}.
Table~\ref{tab:training-stages} summarizes the adapter-training stages.

\begin{table*}[t]
\centering
\scriptsize
\begin{threeparttable}
\caption{Sequential C-WFDM adapter-training stages.}
\label{tab:training-stages}
\setlength{\tabcolsep}{5pt}
\renewcommand{\arraystretch}{1.15}
\begin{tabularx}{\textwidth}{@{}c c l X l@{}}
\toprule
Stage & Target $\tau$ & Required real scans & Conditioning tuple $\mathcal{C}_{\tau}$ & Generated conditioning intermediates \\
\midrule
A & $\mathrm{T2w}$ & T1w, T2w & $(\mathrm{T1w})$ & None \\
B & $\mathrm{FLAIR}$ & T1w, FLAIR & $(\mathrm{T1w},\mathrm{T2w})$ & T2w \\
C & $\mathrm{T1Gd}$ & T1w, T1Gd & $(\mathrm{T1w},\mathrm{T2w},\mathrm{FLAIR})$ & T2w, FLAIR \\
\bottomrule
\end{tabularx}
\begin{tablenotes}[flushleft]\footnotesize
\item[] C-WFDM training stages for T2w, FLAIR, and T1Gd completion. Each stage is supervised with acquired T1w and target scans, while generated T2w and FLAIR intermediates from earlier stages may be used as conditioning inputs.
\end{tablenotes}
\end{threeparttable}
\end{table*}

\subsection{Inference and evaluation protocol}
\label{subsec:evaluation-protocol}

During inference, C-WFDM follows the sequential completion policy from Section~\ref{subsubsec:sequential-completion-policy}, using acquired upstream modalities whenever they are available and synthetic upstream modalities when required by missing intermediate inputs (Figure~\ref{fig:cwfdm-architecture}B).
The generated clean target latent is decoded by the pretrained WF-VAE decoder after inverse WFDM latent scaling, as described in Section~\ref{subsubsec:wfdm-latent-scaling}.

C-WFDM can synthesize missing target modalities for dataset completion even when no acquired target is available.
However, paired SSIM, PSNR, and MAE require an acquired target image as a reference.
Therefore, paired quantitative evaluation for a target modality is restricted to entries with both an acquired T1w input and an acquired target modality.

\subsection{Evaluation metrics}
\label{subsec:metrics}

For all paired target images, we report SSIM, PSNR, and MAE.
SSIM measures structural similarity and is higher when the synthetic image is structurally closer to the real target.
PSNR measures reconstruction quality in decibels, with higher values indicating lower reconstruction error.
MAE measures absolute voxelwise error and is lower when the synthetic image is closer to the paired target.
Metrics are reported as mean $\pm$ standard deviation across evaluated entries.

\FloatBarrier
\section{Experiments}
\label{sec:experiments}

\subsection{Experiment A: BrainScape real-world completion}
\label{subsec:exp1}

This experiment evaluates C-WFDM under the natural modality-availability pattern observed in BrainScape.
The evaluation includes test split entries with an acquired T1w scan because C-WFDM starts from T1w in the sequential completion policy.
No acquired MRI contrasts are deliberately withheld to create incomplete inputs.
For each eligible entry, C-WFDM follows the fixed sequential completion policy defined in Section~\ref{subsubsec:sequential-completion-policy}.
At each completion step, the model uses the acquired conditioning modalities whenever available.
Generated intermediate modalities are used only when a required conditioning modality is missing.
SSIM, PSNR, and MAE are computed only for target modalities that were acquired in the original BrainScape entry.
The target-specific evaluation subsets are determined by the BrainScape test-set modality combinations summarized in Table~\ref{tab:modality-combinations}.

\subsection{Experiment B: BraTS modality completion}
\label{subsec:exp2}

This experiment evaluates C-WFDM on the Brain Tumor Segmentation (BraTS) subset of BrainScape.
Each evaluated BraTS entry includes acquired T1w, T2w, FLAIR, and T1Gd scans.
For each target modality, the corresponding acquired BraTS volume is withheld from the model and used only as the reference for metric computation.
C-WFDM follows the same sequential conditioning policy as Experiment A, with all required conditioning modalities available as acquired scans.
The baseline models use their respective conditioning configurations, with cWDM using all three acquired non-target modalities.

\subsection{Experiment C: BraTS sequential modality completion from T1w}
\label{subsec:exp3}

This experiment evaluates C-WFDM on the same BraTS test subset as Experiment B, but only acquired T1w is provided as input at inference time.
The acquired T2w, FLAIR, and T1Gd scans are retained only as reference targets for metric computation.
C-WFDM generates T2w from acquired T1w, then generates FLAIR using acquired T1w and generated T2w, and finally generates T1Gd using acquired T1w with generated T2w and generated FLAIR.
The acquired T2w, FLAIR, and T1Gd volumes are not used as conditioning inputs in this experiment.
For comparison, ALDM uses the same acquired T1w scan to generate each target directly, without using generated intermediate contrasts.
Metrics are computed for BraTS test participants by comparing each generated modality with the corresponding acquired reference modality.
For C-WFDM, this experiment tests whether the sequential completion pipeline remains stable when later target modalities depend on generated upstream contrasts rather than acquired conditioning inputs.

\subsection{Experiment D: Inference-time efficiency benchmark}
\label{subsec:exp4}

This experiment evaluates the inference time and peak GPU memory usage of C-WFDM and the baseline models.
All benchmarks were performed on a single NVIDIA A100 GPU with 80 GB of memory on the New Zealand eScience Infrastructure (NeSI), using 4 CPU cores and 64 GB of host memory.
The software environment included CUDA 12.6.3, cuDNN 9.5.1.17, Python 3.11.13, and PyTorch 2.4.1.
Inference was performed with a batch size of 1, excluding data loading and disk I/O from timing measurements.
Each configuration was evaluated using 3 warm-up runs followed by 30 timed runs.
We report the mean, median, and standard deviation of inference time, along with peak GPU memory usage.
C-WFDM, ALDM, and cWDM were benchmarked for T2w synthesis using their respective BraTS conditioning protocols.
Since C-WFDM uses separate target-specific adapters, we also evaluated the inference time for FLAIR and T1Gd generation.

\FloatBarrier
\section{Results}
\label{sec:results}

\subsection{Experiment A: BrainScape real-world completion results}
\label{subsec:case1-results}

Table~\ref{tab:case1-results} reports BrainScape modality-completion results using the acquired modalities present in the BrainScape test split.
C-WFDM performs best among the evaluated models for both T2w and FLAIR synthesis.
It reaches SSIM $0.9232$, PSNR $25.80$, and MAE $0.01717$ for T2w, and SSIM $0.8904$, PSNR $24.51$, and MAE $0.02080$ for FLAIR.
For T1Gd, cWDM achieves the best metrics, while C-WFDM is weaker on this contrast-enhanced target.
These results identify T1Gd as the most difficult BrainScape completion target for C-WFDM.

\begin{table}[H]
\centering
\scriptsize
\begin{threeparttable}
\caption{Experiment A: BrainScape real-world completion metrics for T2w, FLAIR, and T1Gd.}
\label{tab:case1-results}
\setlength{\tabcolsep}{4pt}
\renewcommand{\arraystretch}{1.12}
\begin{tabular}{@{}llccc@{}}
\toprule
Target & Model & SSIM $\uparrow$ & PSNR $\uparrow$ & MAE $\downarrow$ \\
\midrule
T2w & ALDM & $0.8928\pm0.0317$ & $22.59\pm2.61$ & $0.02758\pm0.01197$ \\
T2w & \textbf{C-WFDM} & $\mathbf{0.9232\pm0.0328}$ & $\mathbf{25.80\pm2.13}$ & $\mathbf{0.01717\pm0.00704}$ \\
T2w & cWDM & $0.7788\pm0.0279$ & $20.07\pm1.33$ & $0.03757\pm0.00797$ \\
T2w & Pix2Pix & $0.8400\pm0.0306$ & $18.84\pm2.10$ & $0.04412\pm0.01630$ \\
\midrule
FLAIR & ALDM & $0.8517\pm0.0233$ & $21.24\pm1.97$ & $0.03265\pm0.00880$ \\
FLAIR & \textbf{C-WFDM} & $\mathbf{0.8904\pm0.0302}$ & $\mathbf{24.51\pm1.99}$ & $\mathbf{0.02080\pm0.00649}$ \\
FLAIR & cWDM & $0.8415\pm0.0330$ & $20.35\pm2.69$ & $0.03838\pm0.01179$ \\
FLAIR & Pix2Pix & $0.8547\pm0.0286$ & $21.86\pm2.04$ & $0.03126\pm0.00998$ \\
\midrule
T1Gd & ALDM & $0.8668\pm0.0267$ & $22.39\pm1.89$ & $0.02764\pm0.01015$ \\
T1Gd & C-WFDM & $0.8605\pm0.0264$ & $21.48\pm2.46$ & $0.03384\pm0.01409$ \\
T1Gd & \textbf{cWDM} & $\mathbf{0.8907\pm0.0199}$ & $\mathbf{23.63\pm1.93}$ & $\mathbf{0.02498\pm0.00849}$ \\
T1Gd & Pix2Pix & $0.8797\pm0.0248$ & $21.42\pm2.42$ & $0.03406\pm0.01427$ \\
\bottomrule
\end{tabular}
\begin{tablenotes}[flushleft]\footnotesize
\item[] Results are reported as mean $\pm$ standard deviation, with the best value for each metric and target modality shown in bold.
\end{tablenotes}
\end{threeparttable}
\end{table}

\begin{figure*}[t]
\centering
\includegraphics[width=\textwidth]{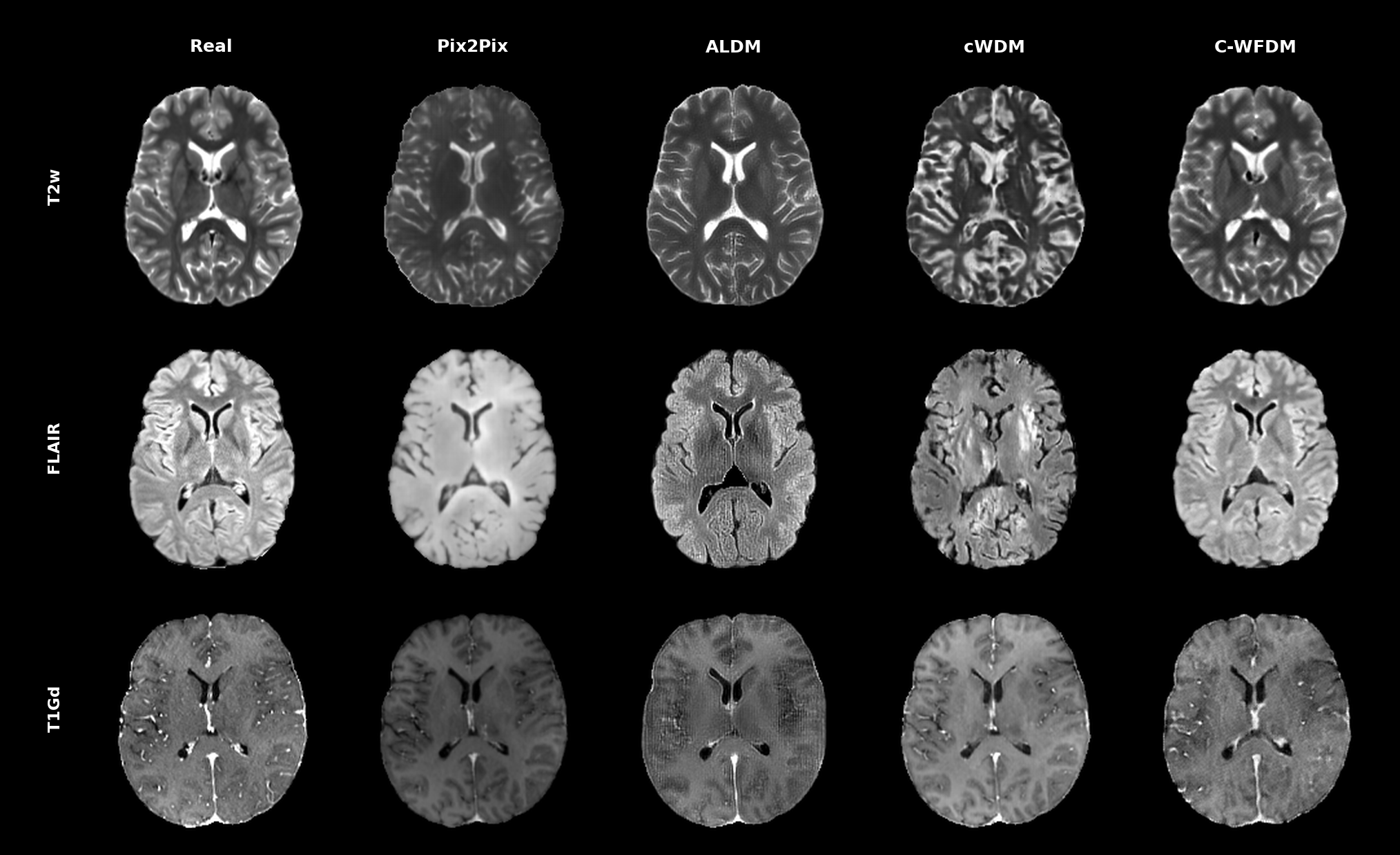}
\caption{Qualitative axial-slice comparison for BrainScape modality completion.
Rows show the T2w, FLAIR, and T1Gd target modalities, and columns show the acquired reference (Real), Pix2Pix, ALDM, cWDM, and C-WFDM.
The T2w and FLAIR examples are from CICT (female, 18 years) and DLBStudy (female, 30 years), respectively, while the T1Gd example is from BraTS.}
\label{fig:brainscape-realworld-qualitative}
\end{figure*}

Figure~\ref{fig:brainscape-realworld-qualitative} compares generated T2w, FLAIR, and T1Gd axial slices with their corresponding acquired images in the BrainScape completion experiment.
For T2w, C-WFDM closely reproduces ventricular anatomy, cortical structure, and tissue contrast, while Pix2Pix shows greater smoothing, ALDM differs in tissue intensities, and cWDM exhibits spatial intensity irregularities.
For FLAIR, C-WFDM preserves ventricular signal suppression and overall anatomical appearance, whereas Pix2Pix shows reduced tissue contrast and ALDM and cWDM exhibit more granular intensity variations.
For T1Gd, C-WFDM preserves the overall anatomy but shows some differences in enhancement detail.
These visual observations are based on three selected BrainScape participants, while the dataset-level results in Table~\ref{tab:case1-results} show that cWDM performs best for T1Gd. 
Additional axial, sagittal, and coronal comparisons are provided in Supplementary Material Appendix~A.

\subsection{Experiment B: BraTS modality completion results}
\label{subsec:case2-results}

Table~\ref{tab:case2-results} reports BraTS modality-completion results.
Every evaluated case includes acquired T1w, T2w, FLAIR, and T1Gd scans, and C-WFDM uses the required acquired scans as conditioning inputs at each completion stage.
For T2w synthesis, ALDM gives the highest SSIM and PSNR, while C-WFDM gives the lowest MAE.
The difference is small, with C-WFDM reaching SSIM $0.8932$ and PSNR $23.92$, compared with ALDM SSIM $0.8940$ and PSNR $23.98$.
This is the most direct ALDM comparison because both models generate T2w from T1w.
For FLAIR, C-WFDM achieves the best scores across all three metrics, with SSIM $0.8734$, PSNR $23.76$, and MAE $0.02376$.
For T1Gd, cWDM gives the best metrics, while C-WFDM remains weaker on this contrast-enhanced target.
Overall, these BraTS results show that C-WFDM is strongest for FLAIR synthesis, remains close to ALDM for T2w, and is less competitive for T1Gd.

\begin{table}[H]
\centering
\scriptsize
\begin{threeparttable}
\caption{Experiment B: BraTS modality completion metrics.}
\label{tab:case2-results}
\setlength{\tabcolsep}{4pt}
\renewcommand{\arraystretch}{1.12}
\begin{tabular}{@{}llccc@{}}
\toprule
Target & Model & SSIM $\uparrow$ & PSNR $\uparrow$ & MAE $\downarrow$ \\
\midrule
T2w & \textbf{ALDM} & $\mathbf{0.8940\pm0.0304}$ & $\mathbf{23.98\pm2.16}$ & $0.02276\pm0.01068$ \\
T2w & C-WFDM & $0.8932\pm0.0308$ & $23.92\pm1.94$ & $\mathbf{0.02253\pm0.00845}$ \\
T2w & cWDM & $0.7646\pm0.0385$ & $20.47\pm1.45$ & $0.03667\pm0.01081$ \\
T2w & Pix2Pix & $0.8295\pm0.0267$ & $18.98\pm1.98$ & $0.04244\pm0.01686$ \\
\midrule
FLAIR & ALDM & $0.8511\pm0.0256$ & $22.85\pm1.74$ & $0.02652\pm0.00732$ \\
FLAIR & \textbf{C-WFDM} & $\mathbf{0.8734\pm0.0235}$ & $\mathbf{23.76\pm1.74}$ & $\mathbf{0.02376\pm0.00677}$ \\
FLAIR & cWDM & $0.8686\pm0.0240$ & $23.22\pm2.32$ & $0.02664\pm0.00887$ \\
FLAIR & Pix2Pix & $0.8556\pm0.0331$ & $21.50\pm2.66$ & $0.03491\pm0.01268$ \\
\midrule
T1Gd & ALDM & $0.8664\pm0.0266$ & $22.39\pm1.89$ & $0.02764\pm0.01015$ \\
T1Gd & C-WFDM & $0.8605\pm0.0264$ & $21.48\pm2.46$ & $0.03384\pm0.01409$ \\
T1Gd & \textbf{cWDM} & $\mathbf{0.8905\pm0.0244}$ & $\mathbf{23.81\pm1.81}$ & $\mathbf{0.02408\pm0.00802}$ \\
T1Gd & Pix2Pix & $0.8797\pm0.0248$ & $21.42\pm2.42$ & $0.03406\pm0.01427$ \\
\bottomrule
\end{tabular}
\begin{tablenotes}[flushleft]\footnotesize
\item[] Results are reported as mean $\pm$ standard deviation, with the best value for each metric and target modality shown in bold.
\end{tablenotes}
\end{threeparttable}
\end{table}

\begin{figure*}[t]
\centering
\includegraphics[width=\textwidth]{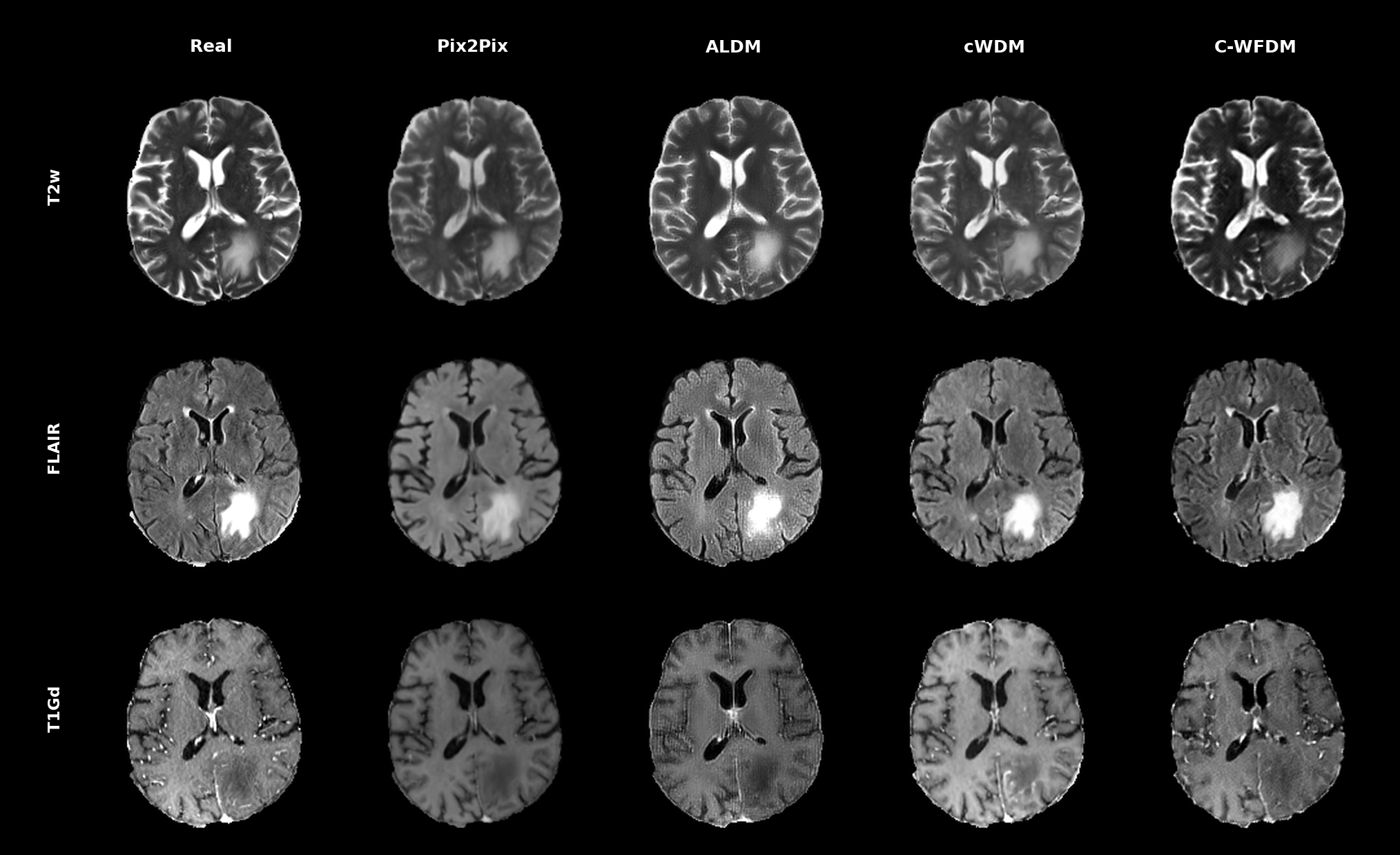}
\caption{Qualitative axial-slice comparison for BraTS modality completion.
Rows show the T2w, FLAIR, and T1Gd target modalities, and columns show the acquired reference (Real), Pix2Pix, ALDM, cWDM, and C-WFDM.}
\label{fig:brats-complete-qualitative}
\end{figure*}

Figure~\ref{fig:brats-complete-qualitative} compares generated T2w, FLAIR, and T1Gd axial slices with their corresponding acquired images in the BraTS modality completion experiment.
For T2w, Pix2Pix appears smoother, while ALDM and C-WFDM preserve ventricular anatomy, cortical structure, and the visible lesion. C-WFDM shows stronger tissue contrast than the reference.
For FLAIR, C-WFDM reproduces the brain lesion, ventricular signal suppression, and surrounding anatomical detail, whereas Pix2Pix appears smoother and ALDM and cWDM show irregular tissue texture.
For T1Gd, C-WFDM preserves the overall anatomy but shows some differences in enhancement detail.
The results in Table~\ref{tab:case2-results} show that C-WFDM performs best for FLAIR and cWDM for T1Gd.
Additional axial, sagittal, and coronal comparisons are provided in Supplementary Material Appendix~B.

\subsection{Experiment C: BraTS sequential modality completion from T1w results}
\label{subsec:case3-results}
Table~\ref{tab:case3-results} reports the BraTS modality-completion results for the T1w-only setting, where only the acquired T1w scan is available for conditioning and the other acquired modalities are used only as reference targets.
In this experiment, C-WFDM first generates T2w from the acquired T1w scan, then generates FLAIR from the acquired T1w scan and generated T2w, and finally generates T1Gd from the acquired T1w scan, generated T2w, and generated FLAIR.
For T2w synthesis, ALDM achieves the highest SSIM and PSNR, while C-WFDM achieves the lowest MAE.
However, the C-WFDM T2w results remain close to ALDM, showing that the proposed model is competitive for T2w synthesis when only T1w is available as the acquired input.
For FLAIR synthesis, C-WFDM achieves the best results across all evaluated metrics, including SSIM, PSNR, and MAE, despite using generated T2w as part of the conditioning input.
For T1Gd synthesis, ALDM achieves the highest SSIM, while C-WFDM achieves the best PSNR and MAE.
These results show that the proposed C-WFDM pipeline remains strongest for FLAIR synthesis and competitive for T2w and T1Gd completion in the T1w-only sequential setting.

\begin{table}[H]
\centering
\scriptsize
\begin{threeparttable}
\caption{Experiment C: BraTS sequential modality completion using acquired T1w.}
\label{tab:case3-results}
\setlength{\tabcolsep}{4pt}
\renewcommand{\arraystretch}{1.12}
\begin{tabular}{@{}llccc@{}}
\toprule
Target & Model & SSIM $\uparrow$ & PSNR $\uparrow$ & MAE $\downarrow$ \\
\midrule
T2w & \textbf{ALDM} & $\mathbf{0.8940\pm0.0304}$ & $\mathbf{23.98\pm2.16}$ & $0.02276\pm0.01068$ \\
T2w & C-WFDM & $0.8907\pm0.0312$ & $23.93\pm2.06$ & $\mathbf{0.02255\pm0.01004}$ \\
T2w & cWDM & $0.7646\pm0.0385$ & $20.47\pm1.45$ & $0.03667\pm0.01081$ \\
T2w & Pix2Pix & $0.8293\pm0.0267$ & $18.97\pm1.98$ & $0.04244\pm0.01686$ \\
\midrule
FLAIR & ALDM & $0.8511\pm0.0256$ & $22.85\pm1.74$ & $0.02652\pm0.00732$ \\
FLAIR & \textbf{C-WFDM} & $\mathbf{0.8554\pm0.0250}$ & $\mathbf{23.10\pm1.68}$ & $\mathbf{0.02558\pm0.00739}$ \\
FLAIR & cWDM & $0.8144\pm0.0210$ & $20.64\pm1.24$ & $0.03479\pm0.00654$ \\
FLAIR & Pix2Pix & $0.8293\pm0.0313$ & $20.50\pm1.95$ & $0.03723\pm0.01108$ \\
\midrule
T1Gd & \textbf{ALDM} & $\mathbf{0.8664\pm0.0266}$ & $22.39\pm1.89$ & $0.02764\pm0.01015$ \\
T1Gd & C-WFDM & $0.8567\pm0.0258$ & $\mathbf{22.69\pm1.76}$ & $\mathbf{0.02647\pm0.00912}$ \\
T1Gd & cWDM & $0.8585\pm0.0266$ & $22.38\pm1.70$ & $0.02868\pm0.00873$ \\
T1Gd & Pix2Pix & $0.8657\pm0.0290$ & $22.40\pm2.04$ & $0.02904\pm0.01072$ \\
\bottomrule
\end{tabular}
\begin{tablenotes}[flushleft]\footnotesize
\item[] Results are reported as mean $\pm$ standard deviation, with the best value for each metric and target modality shown in bold.
\end{tablenotes}
\end{threeparttable}
\end{table}

\begin{figure*}[t]
\centering
\includegraphics[width=\textwidth]{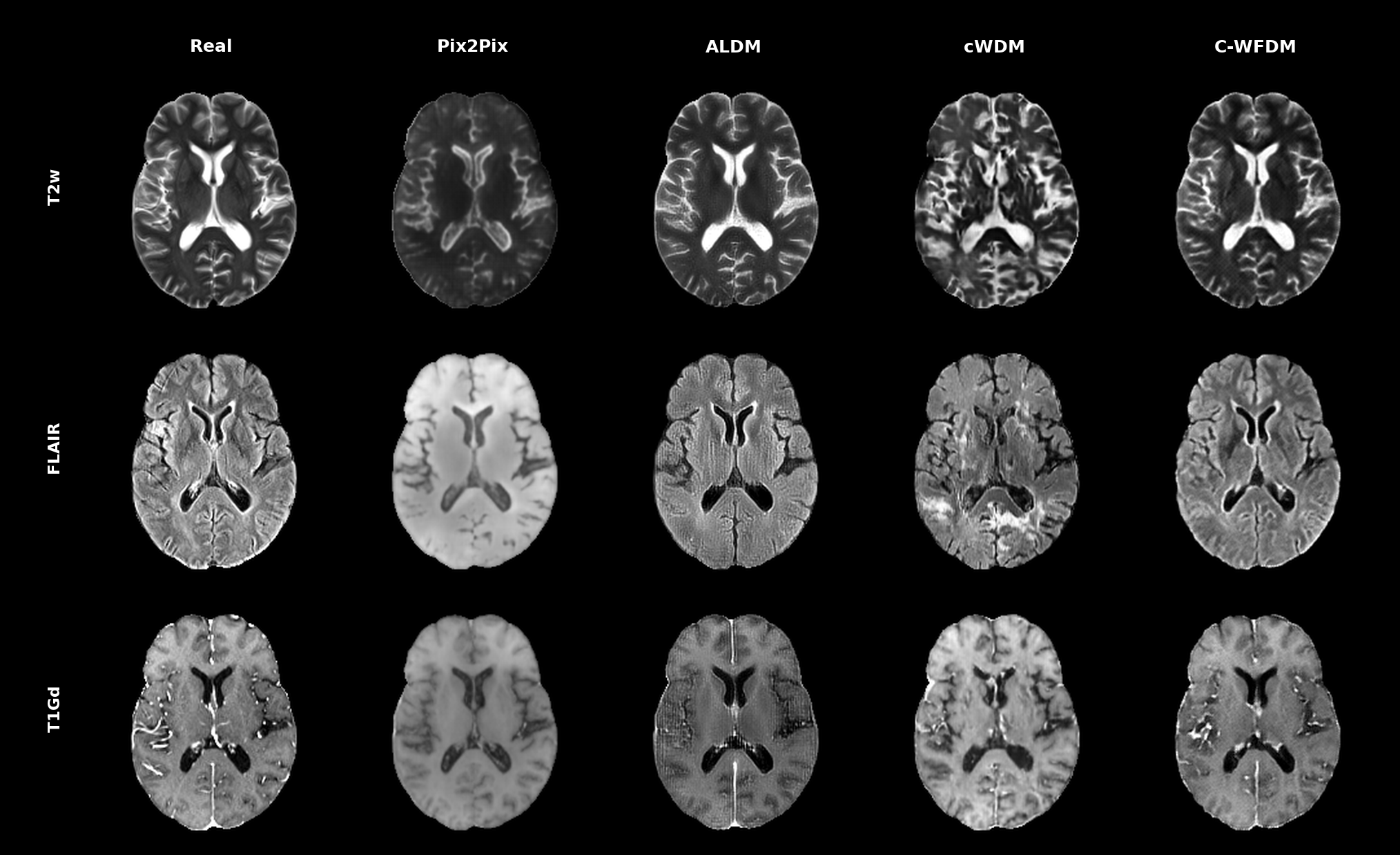}
\caption{Qualitative axial-slice comparison for BraTS sequential modality completion from T1w.
Rows show T2w, FLAIR, and T1Gd, and columns show the acquired reference (Real), Pix2Pix, ALDM, cWDM, and C-WFDM.}
\label{fig:brats-t1w-only-qualitative}
\end{figure*}

Figure~\ref{fig:brats-t1w-only-qualitative} compares generated T2w, FLAIR, and T1Gd axial slices with their corresponding acquired images in the BraTS sequential modality completion experiment starting from T1w.
For T2w, ALDM and C-WFDM preserve brain anatomy and tissue structure, while Pix2Pix appears smoother and cWDM shows brain tissue irregularities.
For FLAIR, C-WFDM preserves ventricular signal suppression and overall anatomy, while Pix2Pix appears smoother, ALDM differs in tissue texture, and cWDM shows greater intensity irregularities.
For T1Gd, C-WFDM preserves the overall anatomy but differs in enhancement detail, while Pix2Pix appears smoother and cWDM shows greater intensity variation.
These visual comparisons complement the quantitative results in Table~\ref{tab:case3-results}.
Additional axial, sagittal, and coronal comparisons are provided in Supplementary Material Appendix~C.

\subsection{Inference-time efficiency benchmark}
\label{subsec:efficiency-results}

Table~\ref{tab:inference-efficiency} summarizes the inference time and peak GPU memory usage of the evaluated models.
We selected T2w synthesis for the cross-model comparison of inference time and GPU memory usage.
ALDM achieved the lowest mean inference time (0.26 s), followed by C-WFDM (3.69 s), Pix2Pix (55.66 s), and cWDM (162.51 s).
Pix2Pix had the lowest peak GPU memory usage (1.14 GB), followed by C-WFDM (3.40 GB).
C-WFDM achieved similar inference times for FLAIR and T1Gd generation (3.60 s each), with peak GPU memory usage of approximately 3.40 GB.

\begin{table}[H]
\centering
\scriptsize
\begin{threeparttable}
\caption{Experiment D: Inference time and peak GPU memory usage.\tnote{$a$}}
\label{tab:inference-efficiency}
\setlength{\tabcolsep}{4pt}
\renewcommand{\arraystretch}{1.12}
\begin{tabular}{@{}llccc@{}}
\toprule
Target & Model & Mean $\pm$ SD (s) & Median (s) & Peak GPU memory (GB) \\
\midrule
T2w & ALDM & $\mathbf{0.2553\pm0.0008}$ & $\mathbf{0.2552}$ & 5.9431 \\
T2w & C-WFDM & $3.6916\pm0.0384$ & 3.6744 & 3.3999 \\
T2w & cWDM & $162.5056\pm0.0910$ & 162.4863 & 5.0502 \\
T2w & Pix2Pix & $55.6596\pm0.5176$ & 55.4910 & $\mathbf{1.1411}$ \\
\midrule
FLAIR & C-WFDM & $3.5987\pm0.0105$ & 3.5990 & 3.3960 \\
T1Gd & C-WFDM & $3.5987\pm0.0096$ & 3.6016 & 3.3992 \\
\bottomrule
\end{tabular}
\begin{tablenotes}[flushleft]\footnotesize
\item[${a}$] Results are reported per generated MRI volume using 3 warm-up runs followed by 30 timed runs for each model. Bold indicates the best T2w result for each metric.
\end{tablenotes}
\end{threeparttable}
\end{table}



\FloatBarrier
\section{Discussion}
\label{sec:discussion}
In this work, we extend WFDM from cohort-level multimodal 3D MRI synthesis to subject-specific missing-modality completion by introducing image conditioning into the BrainScape-trained WFDM backbone.
Target-specific ControlNet adapters use MRI contrasts from the same participant to guide synthesis, while the pretrained WFDM backbone retains its target-modality and metadata conditioning.
The combined model preserves the multimodal synthesis capability of WFDM while adapting it to a more constrained completion setting, where the generated target modality must remain anatomically consistent with the available scans of the same participant.

The sequential completion design reflects the uneven availability of MRI contrasts in BrainScape \parencite{Yasinzai-BrainScape-2025,Yasinzai-WFDM-2026}.
T1w is the most available contrast; T2w and FLAIR are less frequently available, and T1Gd is the rarest, consistent with the selective clinical use of gadolinium-enhanced imaging.
The proposed completion pathway therefore follows a practical pooled-dataset hierarchy from commonly available modalities toward less frequently available and clinically selective target modalities. 
This makes the method different from conventional paired translation settings, where missing modalities are often simulated by removing one contrast from otherwise complete multimodal cases.
In real pooled neuroimaging resources, synthesizing T1w from T1Gd is rarely required and poorly aligned with observed acquisition patterns.

The BrainScape real-world completion experiment provides the most direct evaluation of the intended C-WFDM setting. 
This experiment uses the modality combinations observed in the BrainScape test split. 
C-WFDM achieved the strongest results among the evaluated models for T2w and FLAIR completion (Table~\ref{tab:case1-results}). 
These results support the potential use of C-WFDM for modality completion in pooled neuroimaging resources.

ALDM generates each target from a single source contrast, while cWDM conditions on all three non-target contrasts. C-WFDM instead follows a sequential completion strategy that uses acquired upstream contrasts when available and generated intermediates when required.
In the BraTS evaluation, cWDM achieves the best T1Gd results across the reported metrics when conditioned on all three acquired non-target contrasts (Table~\ref{tab:case2-results}).
Compared with Pix2Pix, C-WFDM uses 3D latent diffusion, the wavelet-fusion representation of the WFDM backbone, and ControlNet-guided sequential multimodal conditioning rather than classical paired adversarial translation.

The three evaluation experiments address complementary aspects of the completion problem.
The BrainScape experiment evaluates the intended heterogeneous pooled-data setting, where modality availability is sparse and uneven.
The BraTS experiment evaluates modality completion with all three non-target MRI contrasts available as acquired scans.
The BraTS T1w-only sequential completion experiment simulates a common scenario in pooled MRI datasets in which T1w is the only acquired conditioning contrast and later targets depend on generated intermediate modalities.

This study has several limitations. First, paired SSIM, PSNR, and MAE evaluation requires an acquired reference target and cannot directly assess entries where the target modality was not acquired. 
Second, sequential completion can accumulate errors. If T2w is generated imperfectly, this error can affect FLAIR synthesis, and errors in both generated T2w and FLAIR can affect T1Gd synthesis.
Although the training strategy exposes later stages to generated intermediate inputs, error propagation remains a central limitation of the sequential design.
Third, C-WFDM is not an arbitrary missing-modality model. The present study follows a BrainScape-motivated completion order and does not exhaustively evaluate every possible input-output modality combination.
Fourth, T1Gd remains a challenging target in this study.
This limitation is consistent with the original WFDM results, where contrast-enhanced synthesis was also more difficult \parencite{Yasinzai-WFDM-2026}.
This difficulty may partly reflect the limited availability of T1Gd in BrainScape.

Future work will evaluate whether C-WFDM-completed modality sets improve downstream AI performance. This includes testing whether completed BraTS modality sets improve tumor segmentation compared with T1w-only or incomplete-modality inputs.

\section{Data availability}
\label{sec:data-availability}

This study uses BrainScape, which aggregates public brain MRI datasets through dataset-specific configuration files and metadata \parencite{Yasinzai-BrainScape-2025}.
Researchers can refer to the BrainScape paper for details on data access, dataset reconstruction, and the preprocessing workflow used to generate the study data from the original public sources.

\section{Code availability}
\label{sec:code-availability}

The code and supporting resources for this study are available through the project repository at \url{https://github.com/yasinzaii/generative-mri}.
The repository serves as a public gateway to the study source code, model configurations, training and sampling workflows, environment specifications, and accompanying software packages.

Reusable model, data-handling, conditioning, inference, and evaluation components developed for this study are released through \textit{BrainMint}, an open-source Python package available from GitHub at \url{https://github.com/yasinzaii/brainmint} \parencite{Yasinzai-BrainMint-2026}.

\FloatBarrier
\section*{Author Contributions}

\begin{enumerate}
  \item \textbf{Muhammad Nabi Yasinzai (MNY)}: Conceptualization, Methodology, Software, Writing -- original draft.
  \item \textbf{Remika Mito (RM)}: Conceptualization, Supervision, Writing -- review \& editing. 
  \item \textbf{Mangor Pedersen (MP)}: Conceptualization, Supervision, Writing -- review \& editing, Resources, Funding acquisition. 
\end{enumerate}
All authors read, revised, and approved the final version of the manuscript.

\section*{Funding}

MNY is a recipient of Australian Epilepsy Project PhD scholarship 
(Medical Research Future Fund - Frontier Health and Medical Research Program - 
Grant Numbers MRFF75908). 
RM is the recipient of an Australian Research Council Discovery Early Career 
Researcher Award (project number: DE240101035). 
MP is a recipient of Health Research Council fellowship, New Zealand (\#21/622).

\section*{Declaration of Competing Interests}

The authors declare that they have no competing financial or non-financial interests 
that could have influenced the work reported in this manuscript.

\section*{Acknowledgements}

We thank the participants in the original studies whose data make this work possible. 
Their contributions to open neuroimaging resources made the BrainScape aggregation possible.

We also acknowledge New Zealand eScience Infrastructure (NeSI) for providing the high-performance computing resources that supported the training and evaluation of the models in this study. 

\FloatBarrier
\section{Supplementary Material}
\label{sec:supplementary}

\subsection{Appendix A: BrainScape completion qualitative examples}
\label{supp:brainscape-qualitative}

\begin{figure*}[p]
\centering
\includegraphics[width=\textwidth]{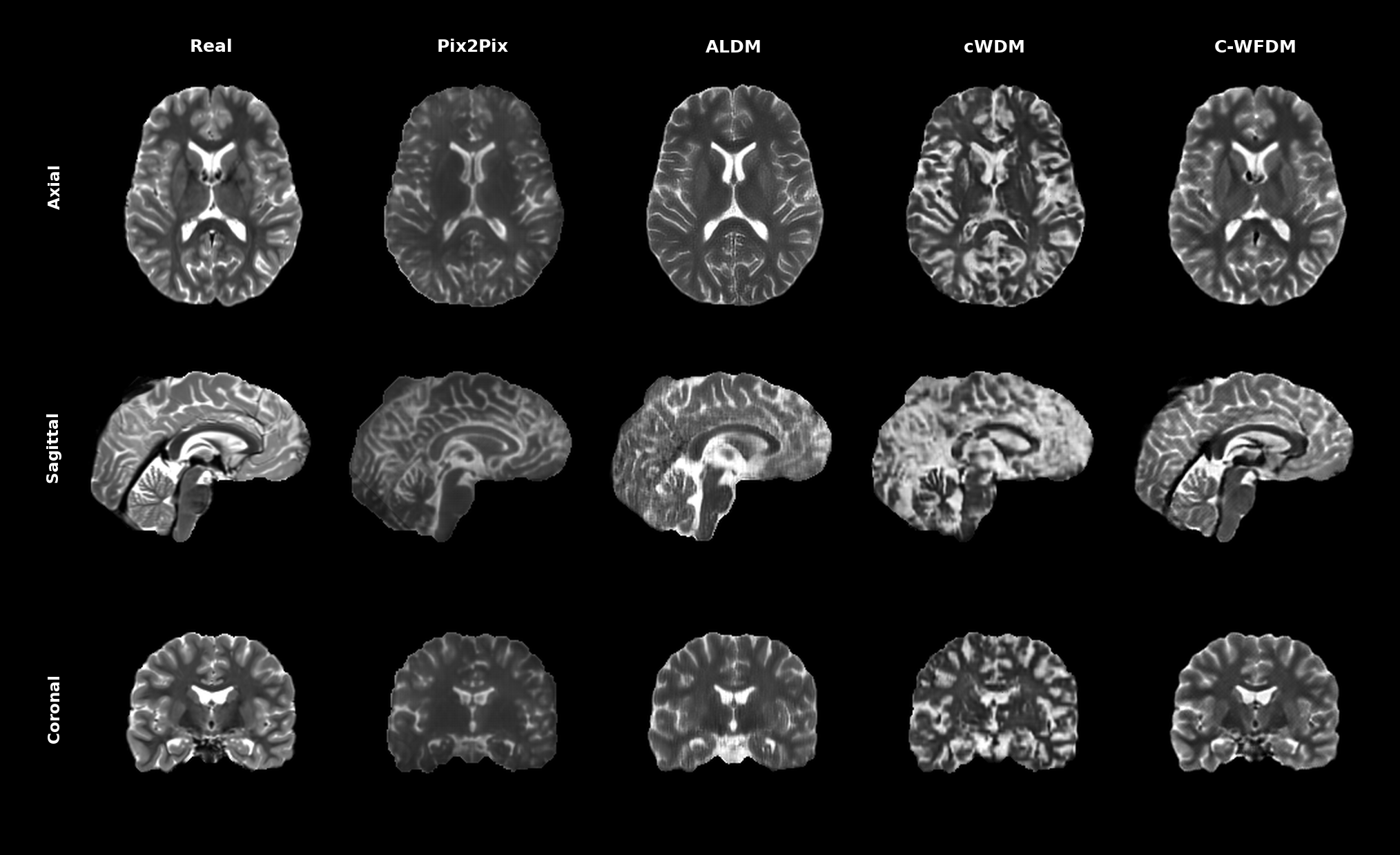}
\caption{Supplementary qualitative comparison of BrainScape T2w completion in axial, sagittal, and coronal views (top to bottom). Columns show Real, Pix2Pix, ALDM, cWDM, and C-WFDM (left to right).}
\label{fig:supp-brainscape-t2w}
\end{figure*}

\begin{figure*}[p]
\centering
\includegraphics[width=\textwidth]{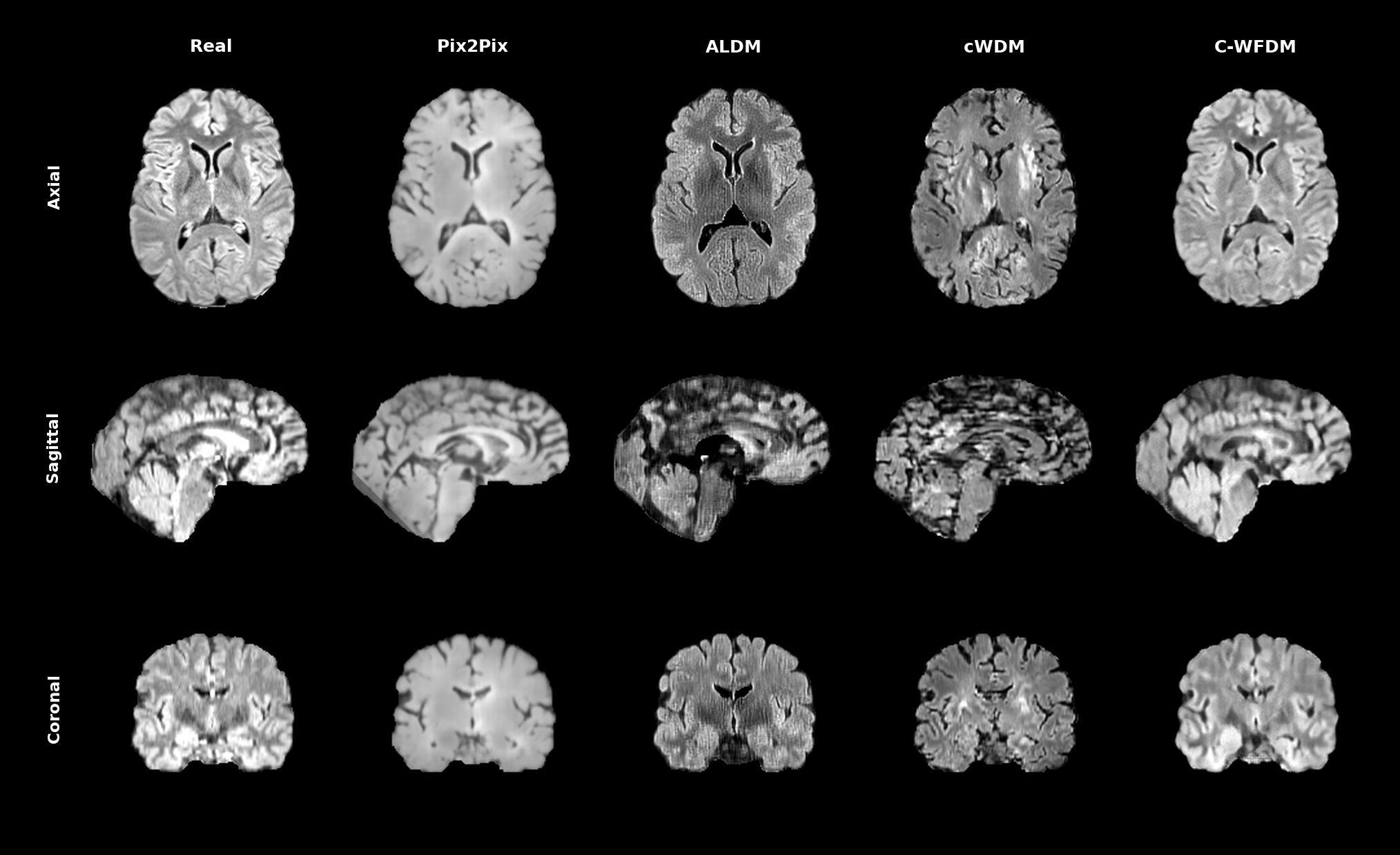}
\caption{Supplementary qualitative comparison of BrainScape FLAIR completion in axial, sagittal, and coronal views (top to bottom). Columns show Real, Pix2Pix, ALDM, cWDM, and C-WFDM (left to right).}
\label{fig:supp-brainscape-flair}
\end{figure*}

\begin{figure*}[p]
\centering
\includegraphics[width=\textwidth]{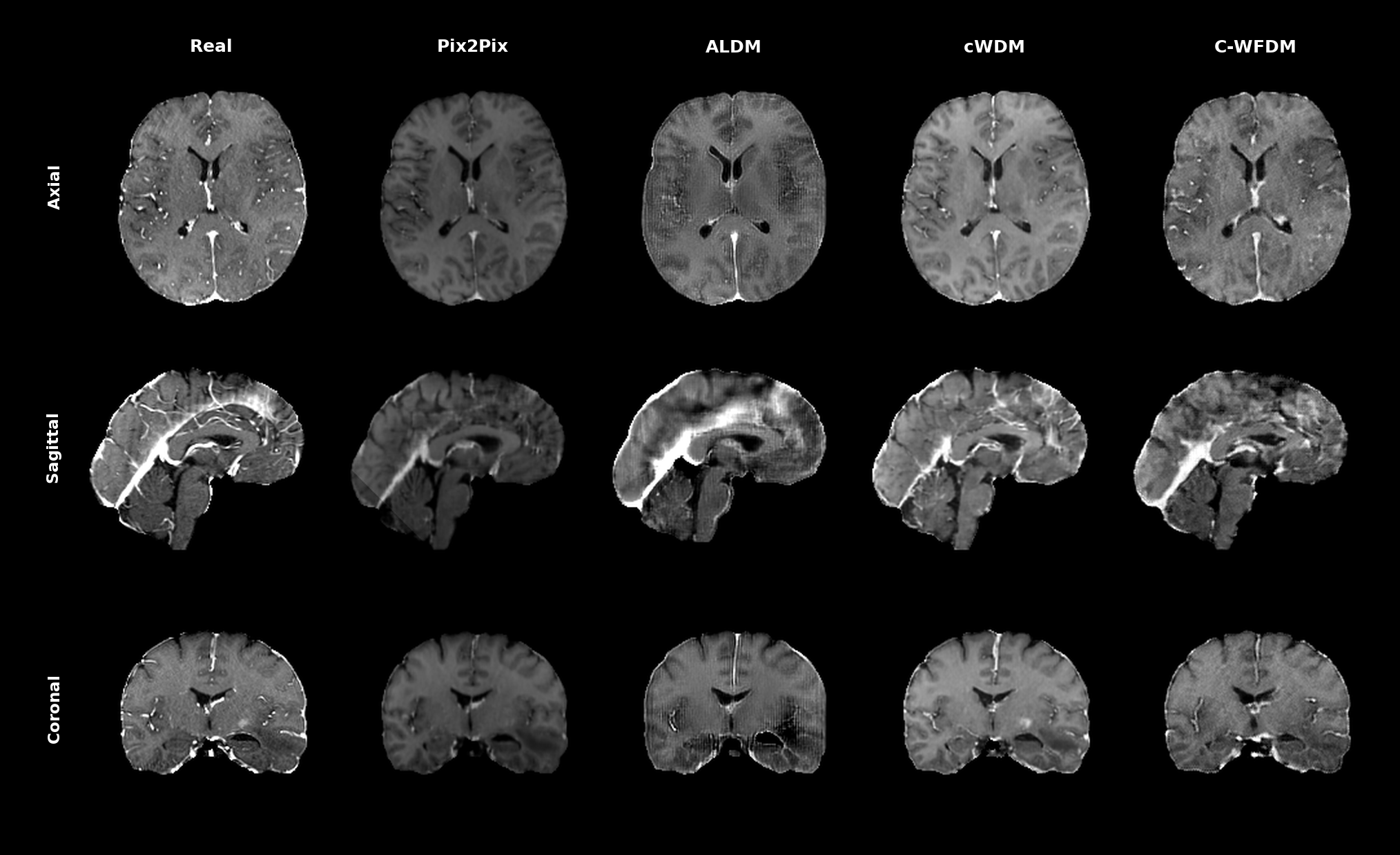}
\caption{Supplementary qualitative comparison of BrainScape T1Gd completion in axial, sagittal, and coronal views (top to bottom). Columns show Real, Pix2Pix, ALDM, cWDM, and C-WFDM (left to right).}
\label{fig:supp-brainscape-t1gd}
\end{figure*}

Figures~\ref{fig:supp-brainscape-t2w}--\ref{fig:supp-brainscape-t1gd} present axial, sagittal, and coronal comparisons of T2w, FLAIR, and T1Gd generation for the same three BrainScape subjects shown in Figure~\ref{fig:brainscape-realworld-qualitative}.

For T2w, C-WFDM preserves ventricular anatomy and cortical structure across all three views, while Pix2Pix appears smoother and ALDM and cWDM show greater intensity differences from the real T2w reference MRI (Figure~\ref{fig:supp-brainscape-t2w}).

For FLAIR, C-WFDM better preserves the real reference's overall tissue appearance and ventricular suppression across orientations, whereas Pix2Pix appears smoother and ALDM and cWDM show intensity irregularities (Figure~\ref{fig:supp-brainscape-flair}).

For T1Gd, all generated outputs differ in enhancement appearance from the acquired reference. C-WFDM preserves the main anatomy, but some fine enhancing structures are less distinct, particularly in the sagittal view (Figure~\ref{fig:supp-brainscape-t1gd}).

\FloatBarrier

\subsection{Appendix B: BraTS modality completion qualitative examples}
\label{supp:brats-complete-qualitative}

\begin{figure*}[p]
\centering
\includegraphics[width=\textwidth]{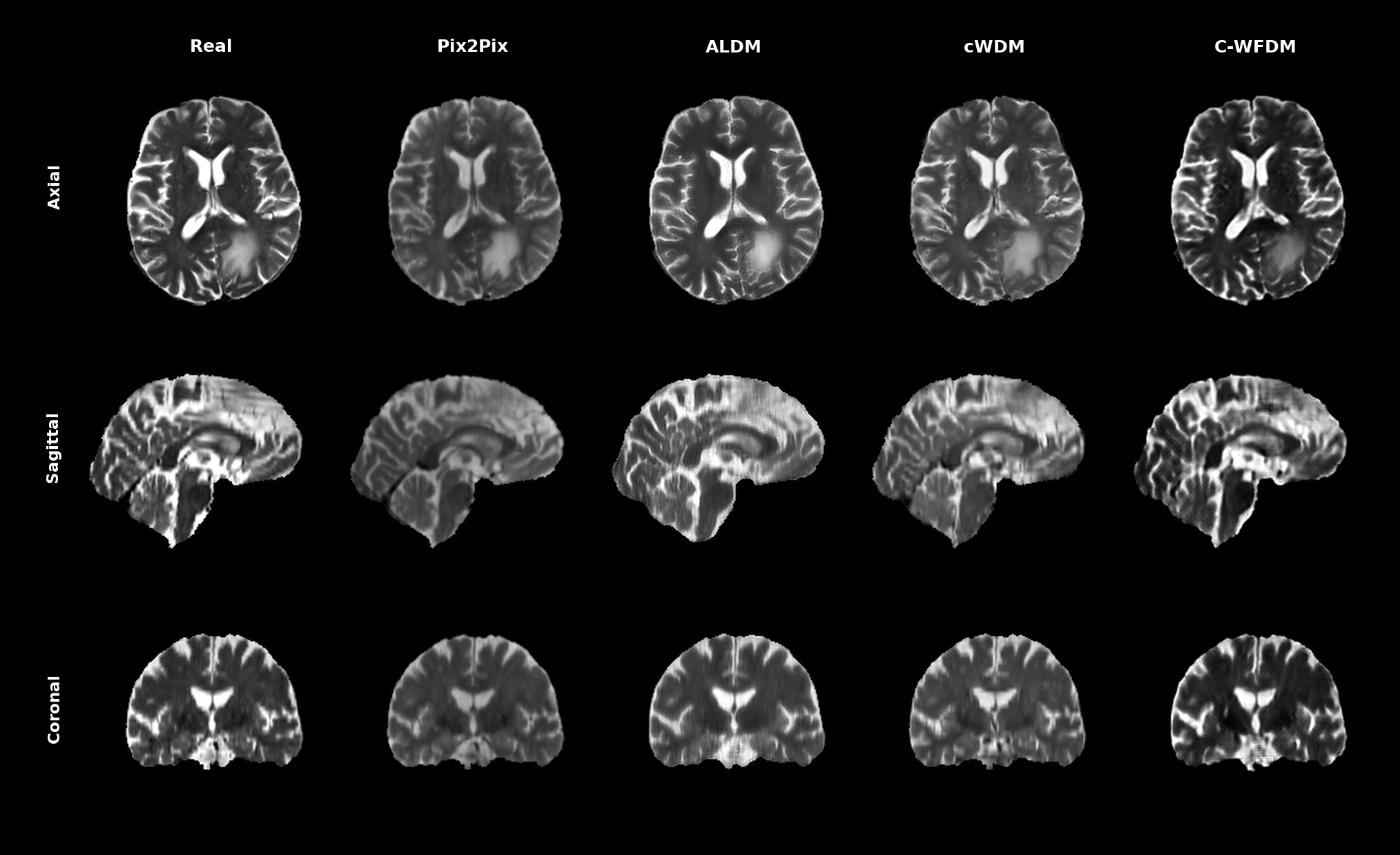}
\caption{Supplementary qualitative comparison of BraTS T2w completion in axial, sagittal, and coronal views (top to bottom). Columns show Real, Pix2Pix, ALDM, cWDM, and C-WFDM (left to right).}
\label{fig:supp-brats-complete-t2w}
\end{figure*}

\begin{figure*}[p]
\centering
\includegraphics[width=\textwidth]{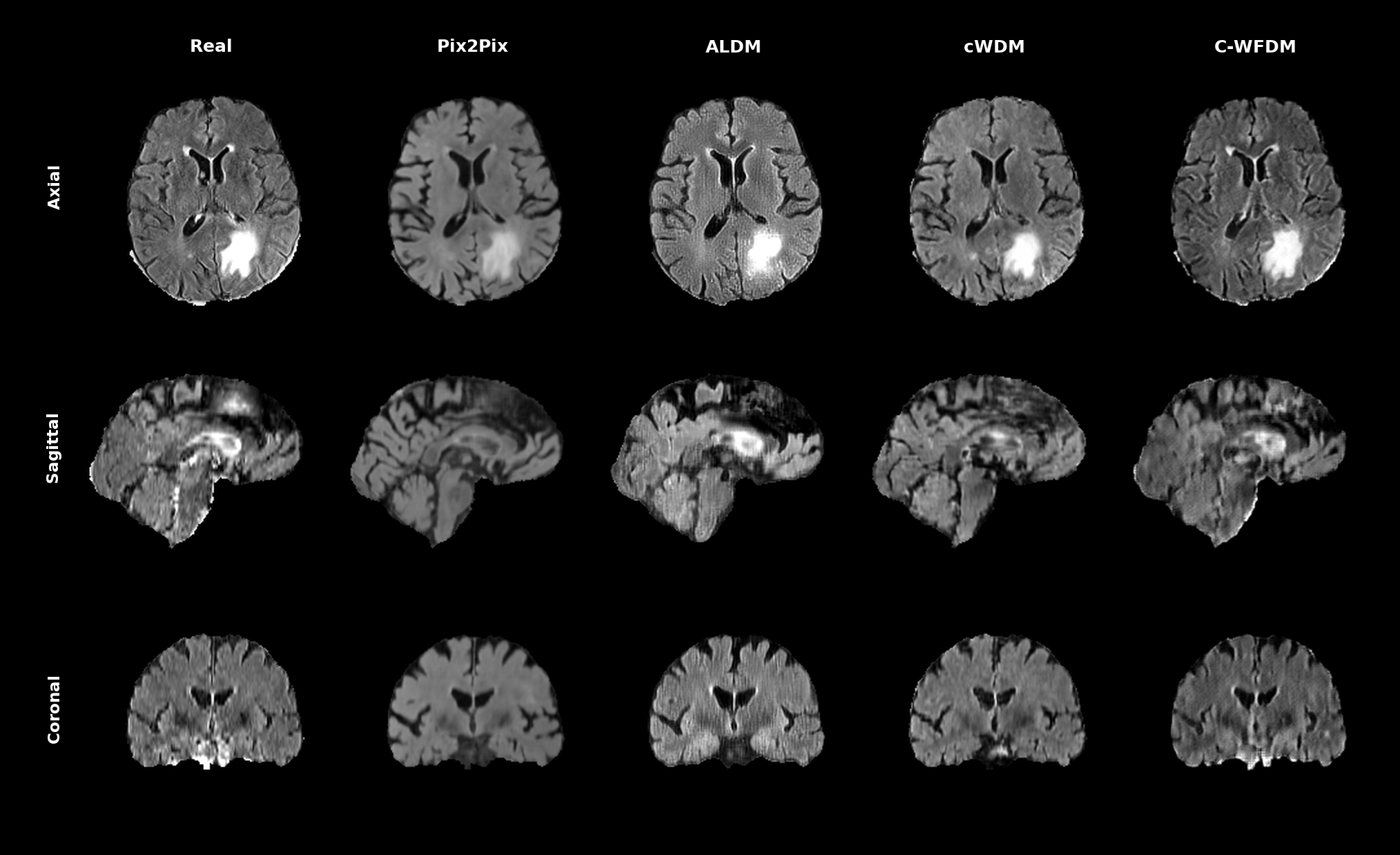}
\caption{Supplementary qualitative comparison of BraTS FLAIR completion in axial, sagittal, and coronal views (top to bottom). Columns show Real, Pix2Pix, ALDM, cWDM, and C-WFDM (left to right).}
\label{fig:supp-brats-complete-flair}
\end{figure*}

\begin{figure*}[p]
\centering
\includegraphics[width=\textwidth]{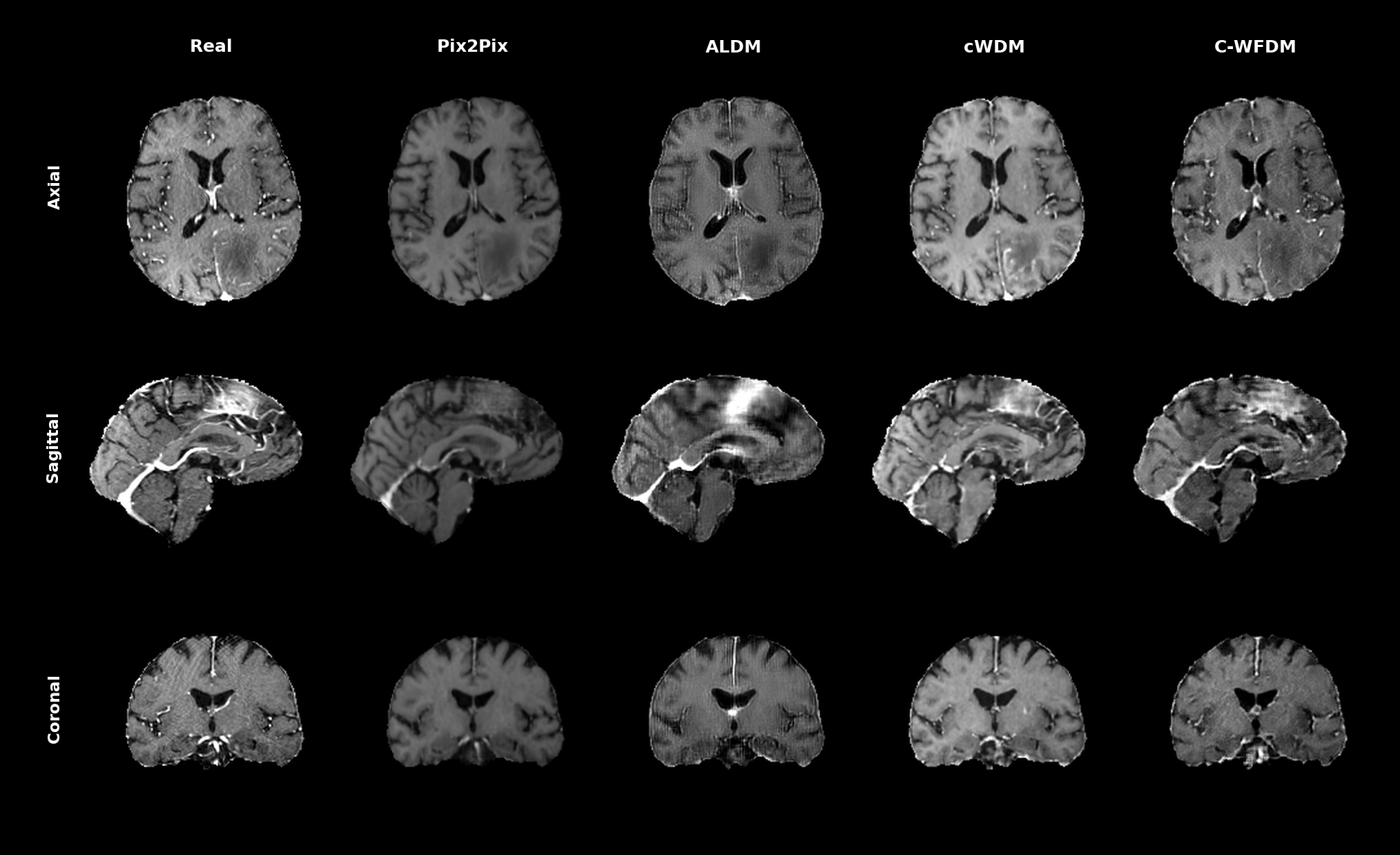}
\caption{Supplementary qualitative comparison of BraTS T1Gd completion in axial, sagittal, and coronal views (top to bottom). Columns show Real, Pix2Pix, ALDM, cWDM, and C-WFDM (left to right).}
\label{fig:supp-brats-complete-t1gd}
\end{figure*}

Figures~\ref{fig:supp-brats-complete-t2w}--\ref{fig:supp-brats-complete-t1gd} show axial, sagittal, and coronal comparisons of T2w, FLAIR, and T1Gd generation for one example from the BraTS modality completion experiment.

For T2w, ALDM and C-WFDM preserve ventricular and cortical structures across all three views, while C-WFDM shows stronger tissue contrast and Pix2Pix appears smoother (Figure~\ref{fig:supp-brats-complete-t2w}).

For FLAIR, C-WFDM preserves ventricular signal suppression, the visible lesion, and overall anatomy, while Pix2Pix appears smoother and ALDM shows greater intensity variation (Figure~\ref{fig:supp-brats-complete-flair}).

For T1Gd, C-WFDM preserves the main anatomy but does not reproduce all enhancement details, particularly in the sagittal view. Pix2Pix appears smoother (Figure~\ref{fig:supp-brats-complete-t1gd}).

\FloatBarrier
\subsection{Appendix C: BraTS sequential modality completion from T1w qualitative examples}
\label{supp:brats-rollout-qualitative}

\begin{figure*}[p]
\centering
\includegraphics[width=\textwidth]{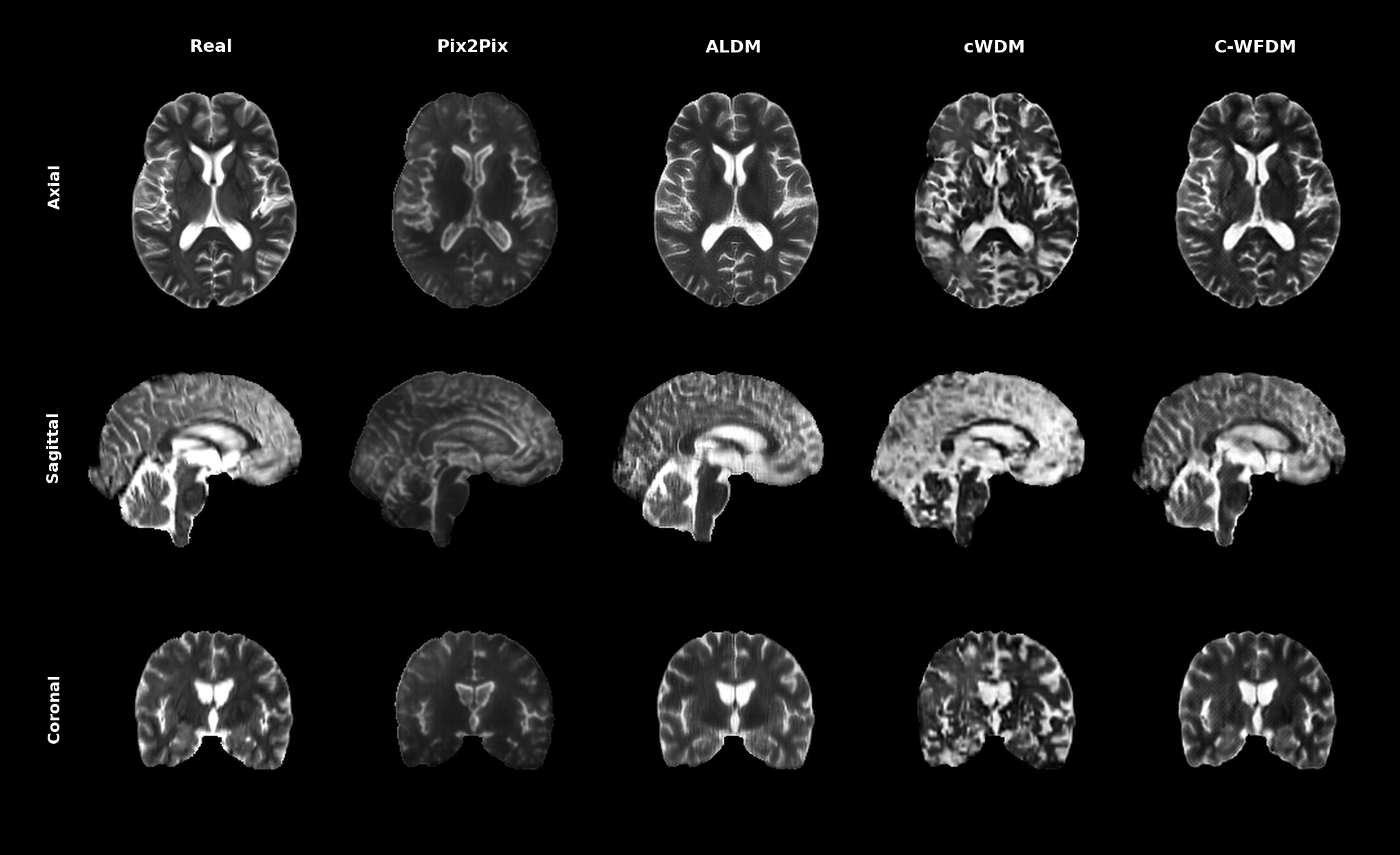}
\caption{Supplementary qualitative comparison of BraTS T1w-only T2w generation in axial, sagittal, and coronal views (top to bottom). Columns show Real, Pix2Pix, ALDM, cWDM, and C-WFDM (left to right).}
\label{fig:supp-brats-rollout-t2w}
\end{figure*}

\begin{figure*}[p]
\centering
\includegraphics[width=\textwidth]{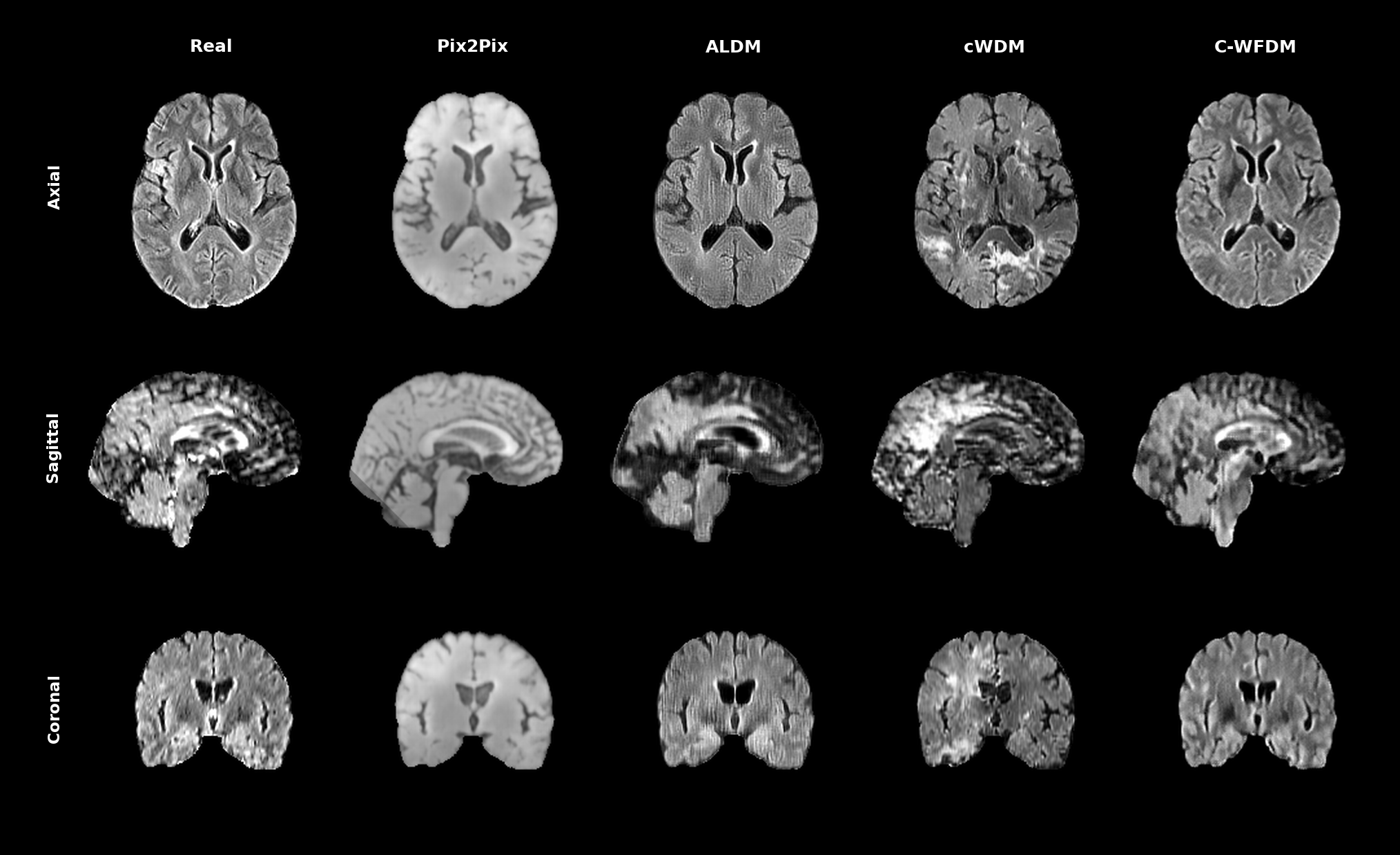}
\caption{Supplementary qualitative comparison of BraTS T1w-only FLAIR generation in axial, sagittal, and coronal views (top to bottom). Columns show Real, Pix2Pix, ALDM, cWDM, and C-WFDM (left to right).}
\label{fig:supp-brats-rollout-flair}
\end{figure*}

\begin{figure*}[p]
\centering
\includegraphics[width=\textwidth]{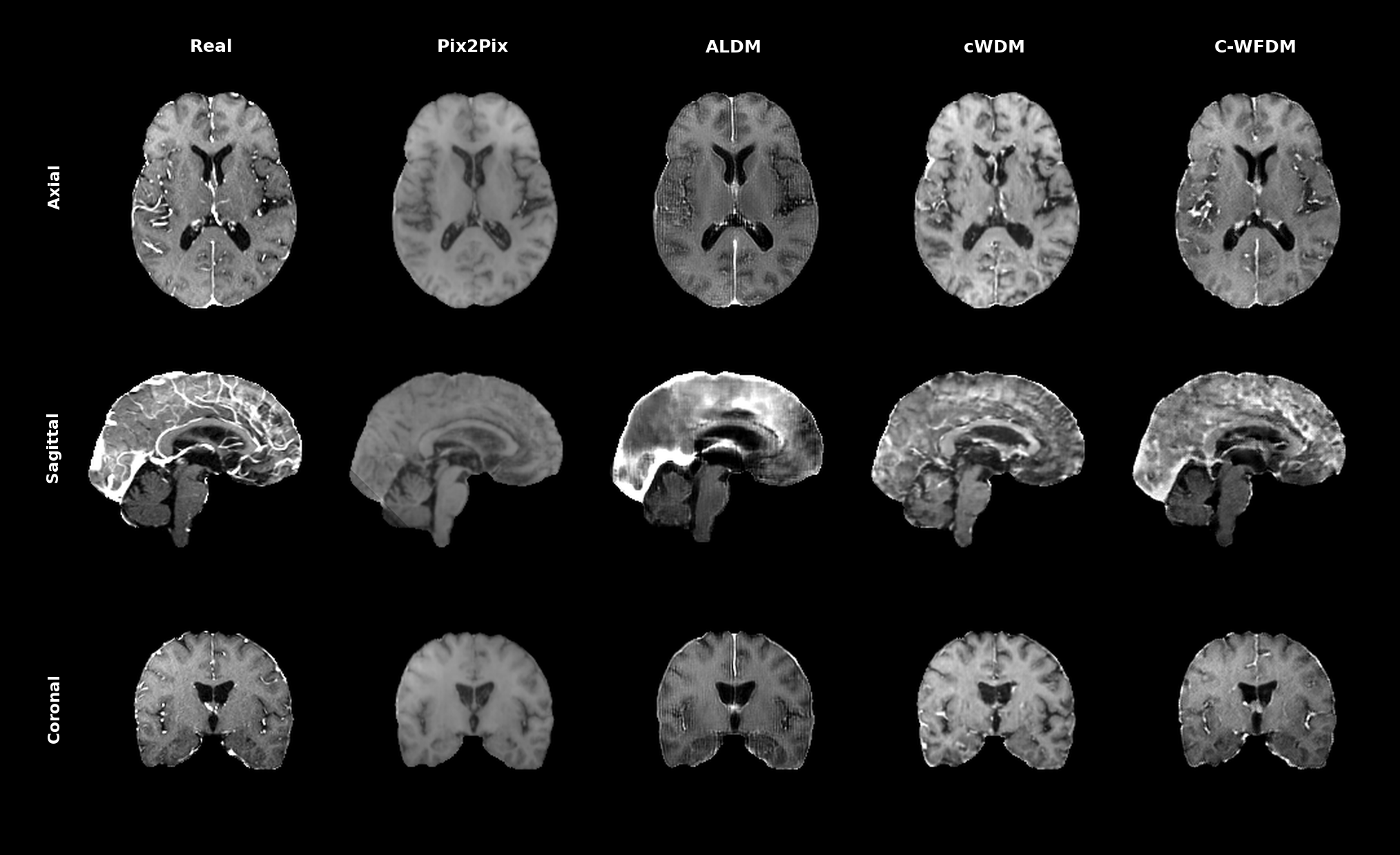}
\caption{Supplementary qualitative comparison of BraTS T1w-only T1Gd generation in axial, sagittal, and coronal views (top to bottom). Columns show Real, Pix2Pix, ALDM, cWDM, and C-WFDM (left to right).}
\label{fig:supp-brats-rollout-t1gd}
\end{figure*}

Figures~\ref{fig:supp-brats-rollout-t2w}--\ref{fig:supp-brats-rollout-t1gd} present axial, sagittal, and coronal comparisons of T2w, FLAIR, and T1Gd generation for the same subject in the BraTS sequential modality completion from T1w experiment.

For T2w, ALDM and C-WFDM preserve ventricular and cortical structures across all three views, while Pix2Pix appears smoother and cWDM shows intensity irregularities (Figure~\ref{fig:supp-brats-rollout-t2w}).

For FLAIR, C-WFDM preserves ventricular signal suppression and anatomy across all three views, while Pix2Pix appears smoother, ALDM differs in tissue texture, and cWDM shows irregular tissue structures (Figure~\ref{fig:supp-brats-rollout-flair}).

For T1Gd, C-WFDM preserves the overall anatomy but shows differences in enhancement detail, particularly in the sagittal view. Pix2Pix shows reduced detail, while ALDM and cWDM differ in regional intensity appearance (Figure~\ref{fig:supp-brats-rollout-t1gd}).

\FloatBarrier
\printbibliography

\end{document}